\documentclass[final]{article}

\usepackage{neurips_2022}

\usepackage[utf8]{inputenc} 
\usepackage[T1]{fontenc}    
\usepackage{hyperref}       
\usepackage{url}            
\usepackage{booktabs}       
\usepackage{amsfonts}       
\usepackage{nicefrac}       
\usepackage{microtype}      
\usepackage{xcolor}         

\usepackage{tikz}
\usetikzlibrary{positioning}
\usetikzlibrary{calc,through,backgrounds}
\usepackage{pgfplots} 
\pgfplotsset{compat=newest}
\usepgfplotslibrary{groupplots}
\usepgfplotslibrary{dateplot}
\usepackage{algpseudocode}
\usepackage{algorithm}
\usepackage[export]{adjustbox}
\usepackage{amsmath,amsfonts,amssymb}
\usepackage{siunitx}
\usepackage{tabularx}
\usepackage{tabulary}
\usepackage{booktabs}
\usepackage{makecell}
\usepackage{multirow}
\usepackage{array}
\usepackage{colortbl}
\usepackage{dcolumn}
\usepackage{stfloats}
\usepackage{xspace,xstring,footmisc}
\usepackage{longtable}
\usepackage[export]{adjustbox}

\definecolor{mycolor}{cmyk}{1,0,1,0}
\definecolor{mycolor1}{rgb}{1.00000,0.50000,0.00000}%
\definecolor{mycolor2}{rgb}{0.00000,0.80000,1.00000}%
\definecolor{mycolor3}{rgb}{1.00000,0.00000,1.00000}%
\definecolor{mycolor4}{rgb}{0.45, 0.31, 0.59}%
\definecolor{mycolor5}{rgb}{0.6, 0.4, 0.8}
\definecolor{carnationpink}{rgb}{1.0, 0.65, 0.79}
\definecolor{auburn}{rgb}{0.43, 0.21, 0.1}

\begin{document}
\let\WriteBookmarks\relax
\def\floatpagepagefraction{1}
\def\textpagefraction{.001}

\title{Thread Counting in Plain Weave for Old Paintings Using Semi-Supervised Regression Deep Learning Models}

\author{%
 A.Delgado \\
  Dep. Teoría de la Señal y Comunicaciones.\\
   ETSI. Universidad de Sevilla.\\
  Camino de los Descubrimientos sn\\
  Sevilla, 41092. Spain \\
   \AND
   Juan.~J.~Murillo-Fuentes\\
  Dep. Teoría de la Señal y Comunicaciones.\\
   ETSI. Universidad de Sevilla.\\
  Camino de los Descubrimientos sn\\
  Sevilla, 41092. Spain 
   \texttt{murillo@us.es} 
     \And
   Laura~Alba-Carcelén. \\
   Dep. Restauración y Documentación Técnica
   Museo Nacional del Prado \\
   Paseo del Prado s/n \\
   28914 Madrid. Spain
}


\maketitle

\begin{abstract}
In this work, the authors develop regression approaches based on deep learning to perform thread density estimation for plain weave canvas analysis. Previous approaches were based on Fourier analysis, which is quite robust for some scenarios but fails in some others, in machine learning tools, that involve pre-labeling of the painting at hand, or the segmentation of thread crossing points, that provides good estimations in all scenarios with no need of pre-labeling. The segmentation approach is time-consuming as the estimation of the densities is performed after locating the crossing points. In this novel proposal, we avoid this step by computing the density of threads directly from the image with a regression deep learning model. We also incorporate some improvements in the initial preprocessing of the input image with an impact on the final error. Several models are proposed and analyzed to retain the best one. Furthermore, we further reduce the density estimation error by introducing a semi-supervised approach. The performance of our novel algorithm is analyzed with works by Ribera, Velázquez, and Poussin where we compare our results to the ones of previous approaches. Finally, the method is put into practice to support the change of authorship or a masterpiece at the Museo del Prado.

\end{abstract}



\maketitle

 \section{Introduction}\label{sec:intro}
\subsection{Interest in Fabric analysis}
In the forensic study of a painting, the analysis of the fabric plays an important role. In particular, the following features are checked \cite{Alba21}:
\begin{itemize}  \setlength\itemsep{.1em}
\item The type of fabric: plain wave, twill, or satin.
\item The fabric material: cotton, linen, silk, ...
\item The number of threads per centimeter in both vertical and horizontal orientation (density of threads), especially in plain weave.
\item The angle deviations of the threads with respect to the horizontal and vertical axis.
\end{itemize}

\begin{figure}[htbp]
\centering
\begin{tabular}{cc}
 \includegraphics[width=3.5cm]{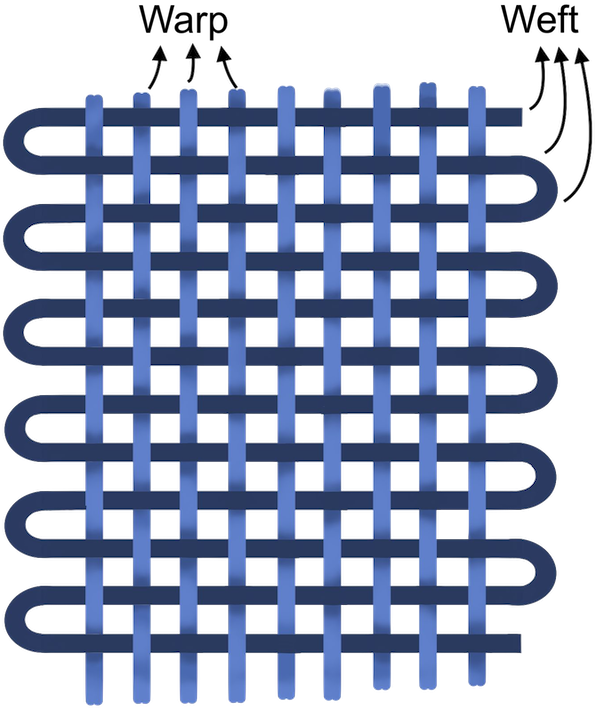} &\includegraphics[width=4.0cm]{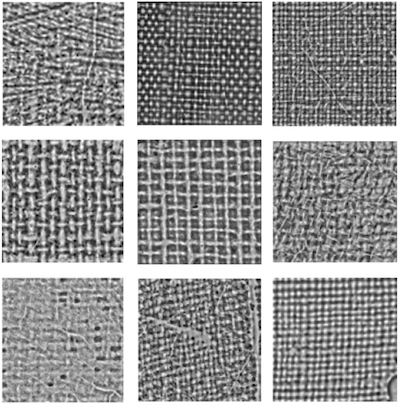}\\
 (a) & (b) 
 \end{tabular}
\caption{In (a) an sketch of plain weave with warp and weft yarns in weaving as described in \cite{Barlow1878} and in (b) patches of 1 cm side from X-ray plates of paintings using plain weave. 
} \label{fig:Taffeta}
\end{figure}

In this work, we focus on the plain weave and its thread density. Plain weave\footnote{Also known as taffeta, tabby, or calico weave.} fabrics are predominant. They are the support for a vast number of paintings thanks to their good compromise between robustness and simplicity \cite{Vanderlip80}. Plain weave is characterized by an intertwining of the vertical threads with the horizontal ones, see Fig. \ref{fig:Taffeta}.a. A set of threads are arranged in parallel on a loom from back to front before starting to weave. These threads are called the warp. The separation between the warp threads follows a deterministic pattern that is determined by their placement on the loom. On the other hand, the weaver passes another thread orthogonally to the warp from one side to the other, tightens it, and passes it back, intertwined. This other thread is the weft. Since weft tightening follows a manual process, 
the distance between weft yarns can be usually modeled by a Gaussian random variable. This crossing of warp and weft threads forms a balanced and robust fabric \cite{Alba21,Simois18}. In fabric analysis with image processing and computer vision, the X-ray of the canvas is used because usually the fabric can not be directly observed as a quite extended technique to reinforce the support of the painting is to stick a piece of cloth on the back. The intensity of each pixel in the image of the X-ray plate depends on the amount of paint and primer in that area, and the presence of wood stretcher, nails, or other opaque artifacts, etc. In Fig. \ref{fig:Taffeta}.b we include some examples of X-ray plates from different canvases using plain weave.


Because the separation of the threads in plain weave is not regular and depends on the manufacturing and the loom, if we find the same pattern in two canvases we can conclude that both come from the same bolt. This is critical to date a canvas or even to change its authorship, by comparison to other well-studied paintings. Traditionally, curators analyzed the fabric of the canvas at some locations, by counting the number of vertical and horizontal threads in 1 cm side squares. However, this provided a rough estimation of the mean value of thread densities. If densities were different enough for a couple of canvases, the curator concluded that they did not share the fabric of a roll. Otherwise, curators could not rule out the possibility of both paintings using the same cloth. With the introduction of automatic thread counting \cite{Johnson2010,Johnson2013} the curators not only avoid the tedious task of counting but they have access to a map of densities for vertical thread throughout the painting, and another for horizontal ones. Angle deviation can also be estimated. By matching these maps, usually depicted using colormaps, the curator concludes on the fabrics of two or more canvases.


\subsection{Motivation and Contributions}

In Fig. \ref{fig:Taffeta}.b we include several examples of patches, of 1 cm sides, to evidence that while counting threads may be an easy task in some cases, it can be really hard in others. Poor resolution, noise from the painting itself or cracks, distortions such as rotations, or a high density of prime make it difficult to count the number of threads in any scenario. We have recently proposed a method for thread density estimation based on deep learning (DL) \cite{AE86,Goodfellow2016} by segmenting crossing points \cite{Bejarano2022a,Bejarano2022b,Bejarano2023}. 
Segmentation of crossing points based on DL presented good results regardless of the impairments present in the input image. However, it is needed the estimation of the densities from the segmentation result, computed in posterior processing. Furthermore, in the segmentation DL paradigm, it is quite difficult to link the segmentation result with the error in posterior thread counting. In this work, we face the direct estimation of the densities by using a regression model. On the one hand, we remove the processing after the segmentation, reducing computational running time. On the other, we use as training loss function the error in the estimation of the thread density, providing a more accurate result. 

One of the major difficulties of the DL is the need for labeled samples in the training stage. This training is run once and then the weights of the model are fixed and ready to be used in the analysis of any painting. A rich and wide set of samples, covering different qualities of fabrics, ranges of thread densities, and noises are needed. But the labeling stage is not only time-consuming but quite hard to accomplish for high-density thread fabrics and noisy images. To create a large dataset, data augmentation (DA) was exploited in \cite{DA2019}, where we cropped areas of labeled samples to generate inputs. Besides, we further augmented the data set by randomly rotating the result of the cropping. Also, the images were pre-processed to get a wide range of intensity values of pixels around a fixed value. Filtering with kernels of fixed size was performed in this step.

We propose a novel algorithm that incorporates three major improvements.
First, we avoid the post-processing of the segmentation result by resorting to regression and forcing the DL model to directly compute the thread density itself. Hence, the output of the network would be the number of vertical threads per cm in that image. Note that to obtain the horizontal density of threads, the input image is rotated 90$^\circ$. We develop and analyze different models to retain the one with the best performance where we perform an optimized search \cite{Snoek12,Omalley19} to set their hyperparameters.

Second, we deeply review the data generation process in \cite{DA2019}. In the pre-processing step, the size of the kernel is not fixed but automatically adapted to the image, depending on a rough estimation of the thread densities. Also, equalization is incorporated. In the DA, we limited the random rotation applied to some of the images and increased the number of images from 21,540 to 30,156 by also cropping central parts of the labeled samples. 

Third, to improve the results in scenarios where the FT provides accurate estimates we introduce semi-supervised training. When processing a new full painting, inputs for which both the DL and the FT provide similar density estimates are incorporated into the training dataset. 

These improvements reduce the error in the estimation of thread densities. Besides, the first one reduces the running time. We include several experiments to compare the outcome of our approach to the one of previous proposals. Finally, we use the new method to report new results on the forensic analysis of a pair of paintings at the Museo del Prado, which helped the curators to change the authorship of one of them.

In summary, the main contributions of the paper are as follows:

\begin{itemize}\setlength\itemsep{.1em}
\item Novel DL models based on regression to directly estimate thread densities for fabrics in old paintings, avoiding cumbersome time-consuming signal processing approaches of previous methods. 
\item Optimized search of the hyperparameters of the proposed models and detailed comparison of the performance of the proposed approaches and the segmentation-based DL method. \cite{Bejarano2023}.
\item Improvements in the DA stage: the limitation of the maximum random rotations and the inclusion of more images from labeled samples to enlarge the dataset.
\item New algorithm to adaptively select the kernel size in the preprocessing approach, adjusting it to the thread densities and enhancing the contrast of the input images. 
\item Using equalization as the last step of the pre-processing to further improve the performance of the approach.
\item A semi-supervised training that automatically includes new labeled samples, those with similar results of the regression DL and the FT approaches.
\item Application of the method to the analysis of masterworks by Velázquez, Poussin, and Ribera at El Museo N. del Prado and The National Gallery, to illustrate the performance of the method compared to the FT, the thread level canvas analysis in \cite{Maaten15} and segmentation DL \cite{Bejarano2023} algorithms.
\item Results on the forensic study that helped to change the authorship of a masterpiece attributed to Rizi at El Museo Nacional del Prado. 

\end{itemize}

\section{Related works}
In the literature, we find three different main techniques for the threads density estimation. Namely, those based on frequency analysis, feature extraction followed by machine learning, and DL.

\subsection{Frequency Analysis}
In \cite{Escofet2001,Johnson2013,Simois18} a theoretical framework is defined in order to model a fabric through a frequency analysis based on the 2D Fourier transform (FT). In the frequency domain, there is a repetition pattern that can be described as a quasi-periodic function in both vertical and horizontal dimensions. The maximum values of this pattern are related to the thread densities of the plain weave. In \cite{Johnson2010,Johnson2013}, this idea was exploited to obtain the first thread counting maps, by applying the 2D discrete FT to 1 cm side square images all over the X-ray plate and finding the maxima. The main advantage of this approach is that it is an unsupervised method, i.e., it does not need any labeling. Besides, it is usually a very robust tool in the presence of noise, such as the painting itself, and artifacts in the X-ray such as nails or stretchers. It also works in poor contrast scenarios. For these reasons, in scenarios where we have a quite uniform warp-weft pattern of threads, the FT provides quite a good result. The FT has been widely used in many studies in the last decade. However, in cases where we do not have a clear and periodic pattern the FT is not useful. In \cite{Bejarano2023} it is reported that the FT fails whenever we have different distances between nearby threads, the width of the threads varies, or the warp is tighter than the weft. In the last case, we clearly observe the warp threads and the weft is observed as a widening of the threads at the crossing points.  



Another interesting work is the one presented in \cite{Simois18} based on the 
power spectral density (PSD). The maximum levels of the PSD exhibit a pattern that characterizes the structure of the fabric, allowing different paintings to be compared. PSD can be viewed as a fingerprint of the canvas as for fabrics with the same thread densities, relevant differences can be observed in the frequency domain. Based on \cite{Simois18}, the Aracne software \cite{Murillo14} was developed. However, it cannot be used to match the warp or weft pattern between canvases.

\subsubsection{Feature Extraction Approach}

Another approach that has been proposed to study the thread density in canvases is based on feature extraction and machine learning \cite{Maaten15}. Compared to the frequency analysis, in \cite{Maaten15} the authors resort to the spatial domain to find every crossing point in the warp-weft pattern. The approach, presented as an automatic thread-level canvas analysis tool, hereafter denoted by ATCA, extracts histograms-of-oriented-gradient features that are the input to machine learning methods such as support vector machine and Bayesian logistic regression classifiers that determines if a location in the images is a crossing point or not. By applying the approach through every pixel, it is possible to generate density maps.


This method exhibits a major drawback, it is necessary to label a large number of crossing points in the images for the to-be-analyzed X-ray plates. This is really cumbersome for the practitioner. This labeling is not trivial, since in many cases threads are not even well observed. Besides, the method depends on a set of parameters whose optimal values depend on the image itself. While some of them are automatically computed, others might need manual adjustment.

\subsubsection{Deep Learning}
Algorithms based on artificial neural networks have been applied to several problems in art. In \cite{Sizyakin20} convolutional neural networks (CNN) were used for crack detection. A CNN was also applied to the automatic classification of paintings \cite{Roberto2020}. In \cite{Pu2020} auto-encoders (AE) were used for image separation. In \cite{Zou21} DL was applied for the virtual restoration of colored paintings. Segmentation through U-Net \cite{Unet15} and AE \cite{AE86,Goodfellow2016} was applied to image restoration by inpainting. 

In the analysis of plain weave fabrics, crossing points can be detected using segmentation based on DL approaches. This is the idea of the method in \cite{Bejarano2022a,Bejarano2022b,Bejarano2023}. A U-Net \cite{Unet15} based model is proposed to locate crossing points in a 1 cm side square input image. Then signal processing is used to estimate the average distance between crossing points. The process is repeated for locations along the canvas to get the density maps. 

In the training stage, a large number of labeled samples were needed. Overall 239 samples of 1.5 cm side images were labeled and preprocessed. Then, 21,540  square images of 1 cm side were generated by cropping these samples and using DA. Although this process is cumbersome, it is performed just once. The curator can later use the trained tool with no additional labeling of the scanned X-ray plate at hand. Hence, compared to the ATCA method, this method avoids any further labeling.
 The preprocessing involves double filtering to reduce the difference in the mean of the input images and increase their contrasts. The kernel size of the involved filters was fixed, regardless of the thread density values. No equalization was applied. 
The algorithm provided good estimates where FT methods fail. In other cases, the FT presented slightly better results.

\section{Data Generation}\label{sec:DG}

The dataset used, the same one as in \cite{Bejarano2023}, has 239 labeled samples of $1.5$ cm side from 36 selected paintings from the Museo Nacional del Prado (MNP). Canvases from Rubens, Velázquez, Lorena, Swanevelt, Dughet, Poussin, Both, Lemaire, and Ribera, among others, were included in the dataset. The fabrics of these paintings have several densities in the range of 6 to 23 threads per cm (thr/cm), different resolutions of the image, and several noise conditions. This encompasses the usual densities found in canvases, see for example the analysis of thread densities in \cite{Vanderlip80} for French painting. 
The models later presented will be trained and tested with this dataset, but the pre-processing and DA are modified as follows.


\subsection{Adaptive Kernel Size}
\begin{figure}[htb]
\centering
\begin{tabular}{cccc}
 \includegraphics[width=3.8cm]{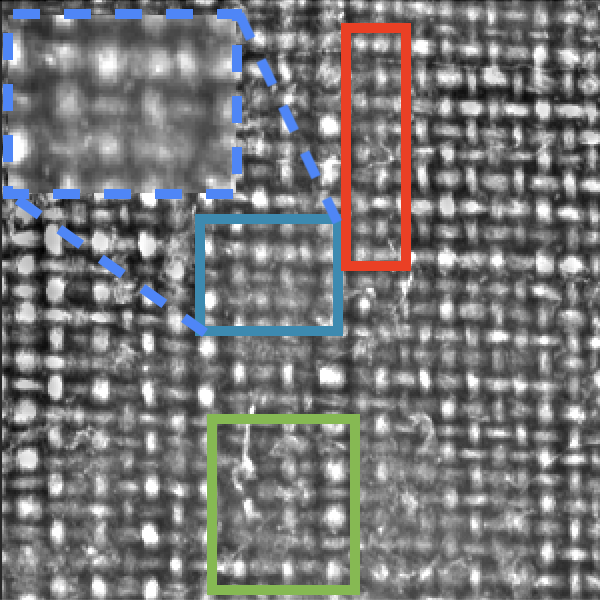}& 
 \includegraphics[width=3.8cm]{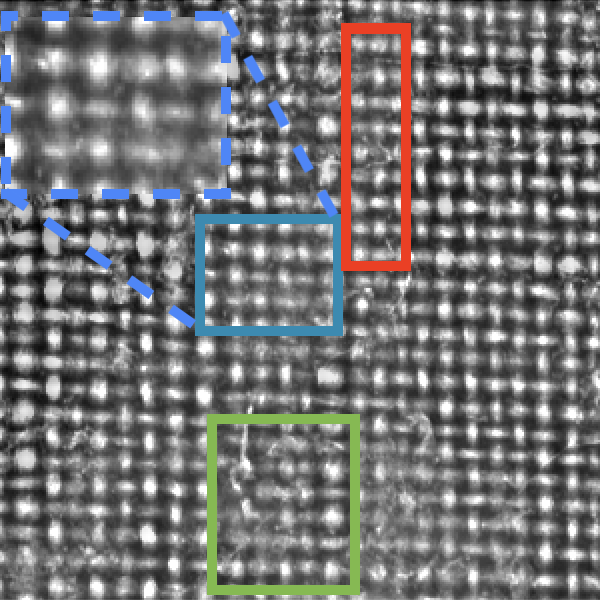}\\
(a) & (b) 
\end{tabular}
\caption{Input image obtained with (a) fixed window size and (b) dynamic window size. We have highlighted some areas and one of them has been zoomed, where after using adaptive kernel size the contrast of the image has been improved. 
} \label{fig:window}
\end{figure}

We have adapted the pre-processing in \cite{Bejarano2023} of the samples from the X-ray plate to better cleanse them. When performing the cropping, mean and local standard deviation filters were applied (see \cite{Bejarano2023} for a full description). Both filters had fixed sizes. We found that images high a better contrast could be obtained by adapting the kernel sizes to the mean fabric thread density. For this reason, we propose to  apply a dynamic window based on the average distance between threads of each fabric, which can be quickly extracted with a coarse frequency analysis of the whole X-ray plate. 
 In Fig. \ref{fig:window}.a we include a pre-processed  sample with fixed size kernel and in Fig. \ref{fig:window}.b we have the same sample pre-processed with adaptive kernel size. We highlighted some areas where a better contrast is observed, i.e., it is easier to distinguish crossing points from dark areas between threads. We also zoomed one of them.
 
 The algorithm to estimate the size of the kernel to pre-process the image is described in Alg. \ref{alg:VW}. This method is run twice, with an initial value $k_0=21$ and then with the result of this first iteration. The linear regression in the algorithm was adjusted to provide the best kernel size from the estimation of the thread densities, based on several previous results processing canvases.

 
\begin{algorithm}
\caption{Kernel Size Estimation}\label{alg:VW}
\textbf{Input:} Scanned X-ray plate image and initial kernel width $k_0$.
\begin{algorithmic}[1]
\State Pre-process the X-ray with kernel width of $k_0$.
\State Rough estimation of the thread densities with FT of 512 points by sampling $1 \times 1$ cm images from the X-ray plate every 7-15 cm, depending on the canvas size.
\State Estimate the histogram for the vertical and horizontal densities, typically with $300$ bins, and retain the largest values above the mode divided by 1.5. Average the horizontal and vertical values, $t$.
\State Estimate the pre-processing kernel size by applying the following linear regression, 
\begin{equation}
k = -0.90 \cdot t + 37.05
\end{equation}
Then $k$ is first rounded to the nearest value to 14.5, and then to the lowest odd number. 
\\
\textbf{Output:} $k$
\end{algorithmic}
\end{algorithm}

\subsection{Equalization}
We include also an off-the-shelf equalization approach. We estimate the histogram, $h(x)$, for the gray levels of the input image, normalize it to sum 255 and compute its integral, $h'(x)$, and use the result as a look-up table. For a given input value, $x$, the output yields $y=h'(x)$.

\subsection{Rotations}
 The fabric of the canvases is usually distorted in some areas. In fact, along the stretchers, the fabric bends due to the tightness around the nails. In other areas of the canvas, the threads are not perfectly arranged vertically or horizontally; in some areas, the deviations can be severe. We include randomly rotated images in the data set to allow the model to learn the densities in rotated scenarios. However, we found that the maximum allowed rotations were larger than needed, and the model parameters were biased towards angle deviations not found in practice.  To avoid this bias we limited the maximum random rotations applied to half the maximum in \cite{Bejarano2023}. Hence, random rotations applied in the DA vary in the ranges [-6$^\circ$, -4$^\circ$], [-3.5$^\circ$, -1$^\circ$], [1$^\circ$, 3.5$^\circ$] and [4$^\circ$, 6$^\circ$]. 
 
\subsection{ Increasing the Dataset }
 The DA has also been modified in order to get more input images for every $1.5 \times 1.5$ cm labeled sample. Previously, in \cite{Bejarano2023} 30 samples were extracted from each sample. These images were mainly taken from areas in the corners of the labeled samples. Now, we add the cropping of the central regions given by the (50:250, 50:250), (65:265, 65:265), (35:235, 35:235), and (80:280, 80:280) pixels. These images are cropped from the original sample and the vertically and horizontally flipped versions (see \cite{Bejarano2023} for further details). In summary, now we get 42 images of $1 \times 1$ cm for every labeled sample, 40\% more than in \cite{Bejarano2023}. No randomly rotated versions of these images are included. Therefore, the ratio of the number of randomly rotated images to the overall number of images decreases. The number of images in the dataset is increased to 30,156.

%
%
%
%

\section{Regression DL Models} \label{sec:DL}

\begin{figure}[htbp]
\centering
 \includegraphics[width=7.7cm]{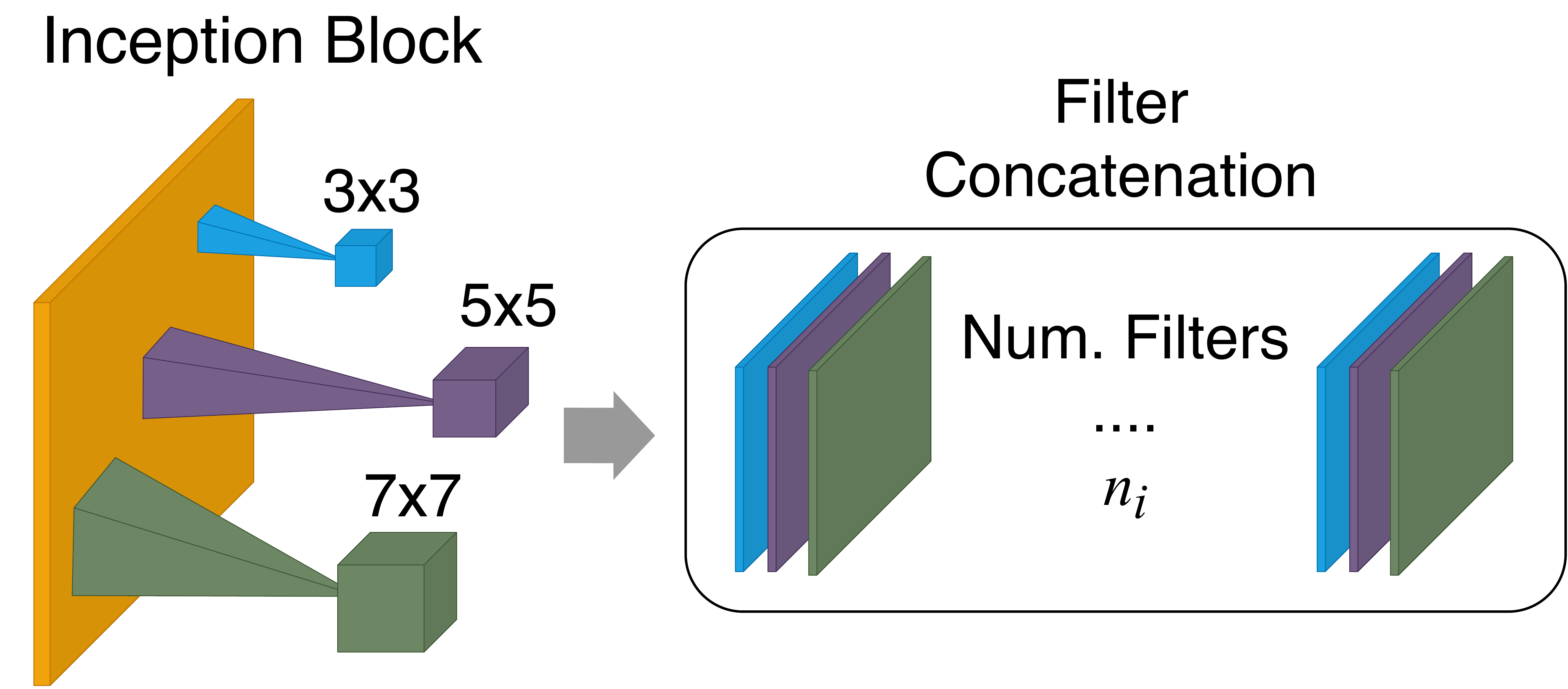}
\caption{Inception block: $3\times3$, $5\times5$, and $7\times7$ kernels are convolved with the input at the same depth, as many times as given by the number of filters parameter, $n_i$. The results of these convolutions are concatenated.} \label{fig:Inception}
\end{figure}

We designed four DL models to perform regression, providing the vertical density of threads per cm. All models follow the inception paradigm as in \cite{Bejarano2023} it was shown to be useful when dealing with images of different densities of threads. By applying the inception paradigm, we have convolutional kernels of several sizes in the same layer:  $3\times 3$, $5\times 5$, and $7\times 7$, see Fig. \ref{fig:Inception}. The results of the convolutions are concatenated at the output. Then, batch normalization is performed followed by a ReLU activation function. This will be used as a central block in the models below, represented by horizontal arrows in the schemes. The number above the horizontal arrows indicates the number of kernels used for each size. Hence the number of resulting features will be this number multiplied by 3. The rectangular prisms indicate the set of features at that point between layers. The number below the prisms are of the form $width \times height \times features$. All vertical arrows but the last two of them at the bottom denote a 2D max-pooling layer. The penultimate arrow indicates a flattening and the last one denotes a set of dense layers.

\subsection{Labeling}

The data generation follows the procedure in \cite{Bejarano2023} with the modifications described in Section \ref{sec:DG} and an extra step to generate the labels for regression. 
While for the segmentation approach the label is an image with the intersection of the annotations for vertical and horizontal threads, for regression we need the density of vertical threads. The density of horizontal threads will be obtained by using the same model with 90$^\circ$ rotated images. Therefore, in the dataset, every image will be also included after applying a 90$^\circ$ rotation. For all of them, we need to estimate the density of vertical threads. 
To compute the corresponding density of vertical threads for every image, we use the spatial counting (SC) approach developed in \cite{Bejarano2023} to estimate the thread density after the output of the segmentation DL models. At this point, it is important to remark that the SC algorithm is used just in the dataset generation in the training of the models but not later in the processing of a canvas, where the regression DL model directly provides the value of the thread density.

The dataset is divided into three subsets: training, validation, and testing. Instances from the same canvas are included just in one of these subsets. The input test dataset is generated as described in \cite{Bejarano2023}, then duplicated after a 90$^\circ$ rotation. Labels were generated by using SC. 
The test subset is not used in the training stage or to select the model or hyperparameters, but just to analyze the final results.
We next propose four models. We later train and analyze them, to retain the one with the best performance. 

\subsection{Inception U-Net Pre-Trained Encoder}\label{ssec:RegUnet}
\begin{figure}[htbp]
\centering
 \includegraphics[width=7.7cm]{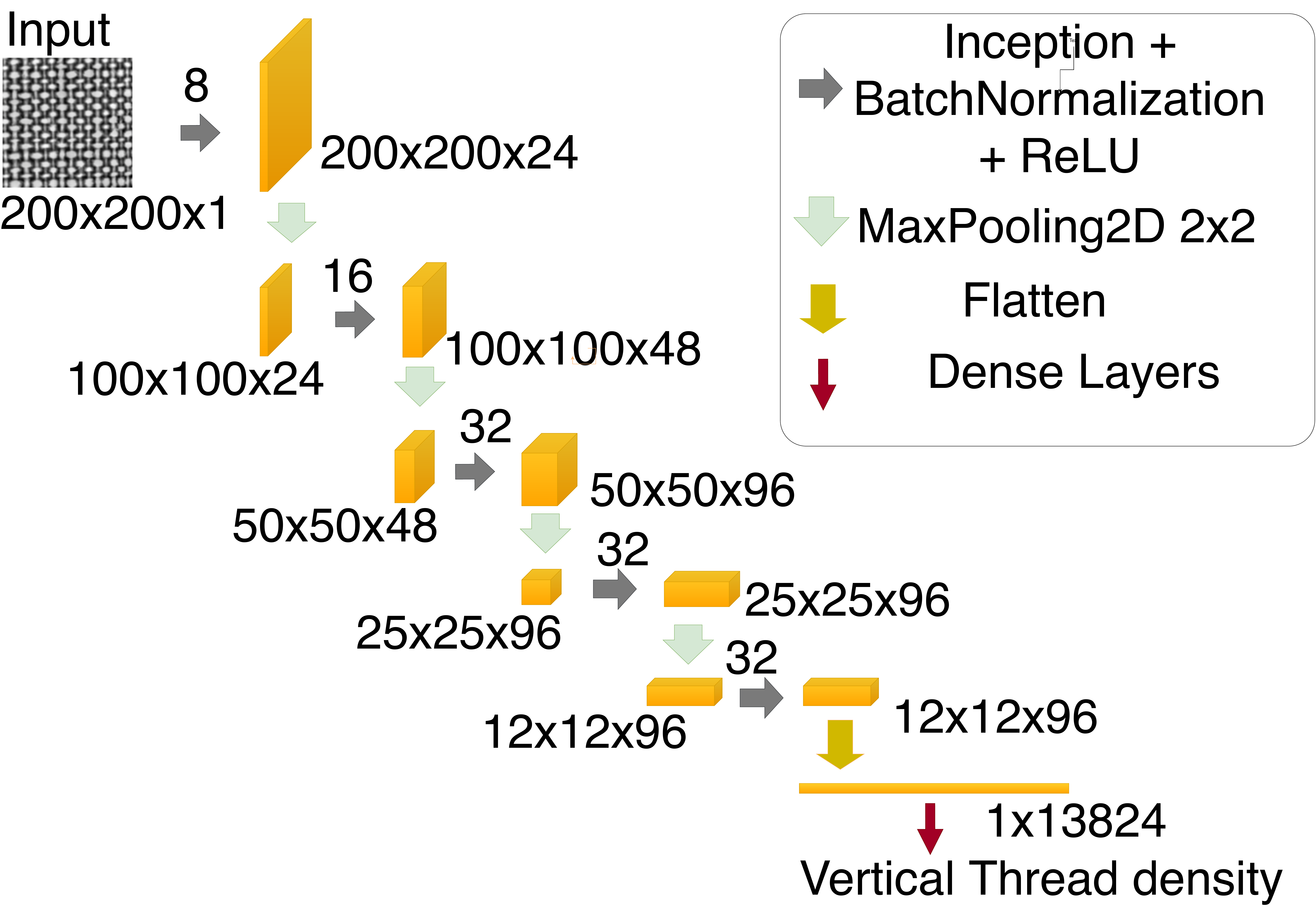}
\caption{Inception U-Net pre-trained encoder adapted to regression task. The number of filters in each convolutional layer is highlighted above each horizontal arrow.} \label{fig:UNET_REG}
\end{figure}

In this model we use a first set of layers corresponding to the pre-trained encoder from the Inc-Dice model in \cite{Bejarano2023}, a model exploiting the U-Net architecture and the inception paradigm. The output of these layers encodes the information to later locate the crossing points in the decoder. But as we need the estimation of the thread densities, we replace the decoder part of the U-Net with a fully connected (FC) network to perform regression, see Fig. \ref{fig:UNET_REG}. 
The number of dense layers and the number of neurons in each layer was selected by using Bayesian optimized search \cite{Snoek12,Omalley19}. 
As a result, we obtained 6 dense layers with 50, 100, 50, 100, 100, and 1 neurons. All layers use ReLU as activation functions except for the output neuron that uses a linear activation to perform the regression. The full model has around 1.38 million parameters, where 0.67 million correspond to the FC layers. 

\subsection{Inception Regression Model}\label{ssec:Reg}

Next, we design a model from scratch, keeping the previous idea of an encoder followed by an FC. Therefore, this architecture has two main parts. The first part performs the feature extraction via convolutional layers. Each convolutional layer has two consecutive inception blocks and 2D max-pooling is applied between layers to reduce dimensionality. Dropout is applied after each pooling. The second part uses a flatten layer
and six dense layers that perform the regression according to the extracted features. ReLU activation function is present in every layer except for the output layer where we use a linear activation. 

\begin{figure}[htbp]
\centering
 \includegraphics[width=8.1cm]{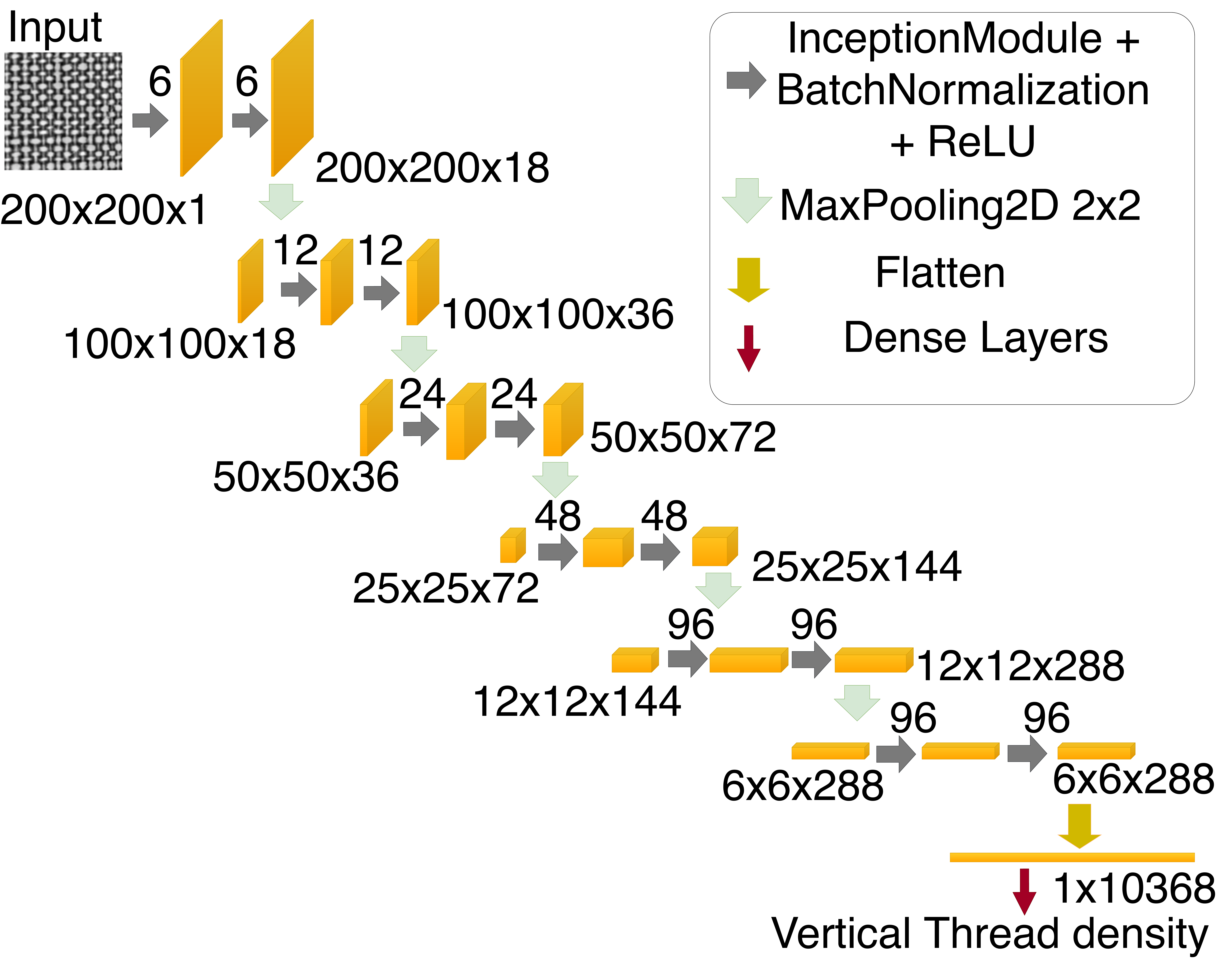}
\caption{Inception regression model.} \label{fig:REG}
\end{figure}

We performed an optimized search \cite{Snoek12,Omalley19} to set the optimal value for the number of convolutional and dense layers, the number of filters in each layer, the number of neurons in each dense layer, and the dropout value. The optimal architecture has an encoder with 6 double CNN layers, i.e., each layer has 2 consecutive Inception blocks. At the output of the encoder, the model includes 5 dense layers with 100, 100, 80, 100, and 1 neurons.  The number of kernels at each CNN layer is described in Fig. \ref{fig:REG}, it doubles as we go down in the encoder, except for the last convolutional layer. The full model has over 10 million parameters.

\subsection{Inception VGG-Based Regression Model}\label{ssec:ResVGG}
In this new architecture, we modify the last model to adapt it to the VGG16 \cite{Simonyan2015} architecture but perform a regression instead of classification. Inception blocks are still used in each convolutional layer. The number of dense layers and the number of neurons in each of them are defined following the VGG architecture. There are two dense layers with 64 times the number of filters of the first layer, $8 \cdot 64$, plus one output layer of size one. 
As in the previous models, an optimized search has been used to set the optimal number of convolutional filters (8 in this case) and the dropout value (0.09). The ReLU activation function is present in every layer except for the output neuron that uses linear activation. The full model has over 9 million parameters.

\begin{figure}[htbp]
\centering
 \includegraphics[width=8cm]{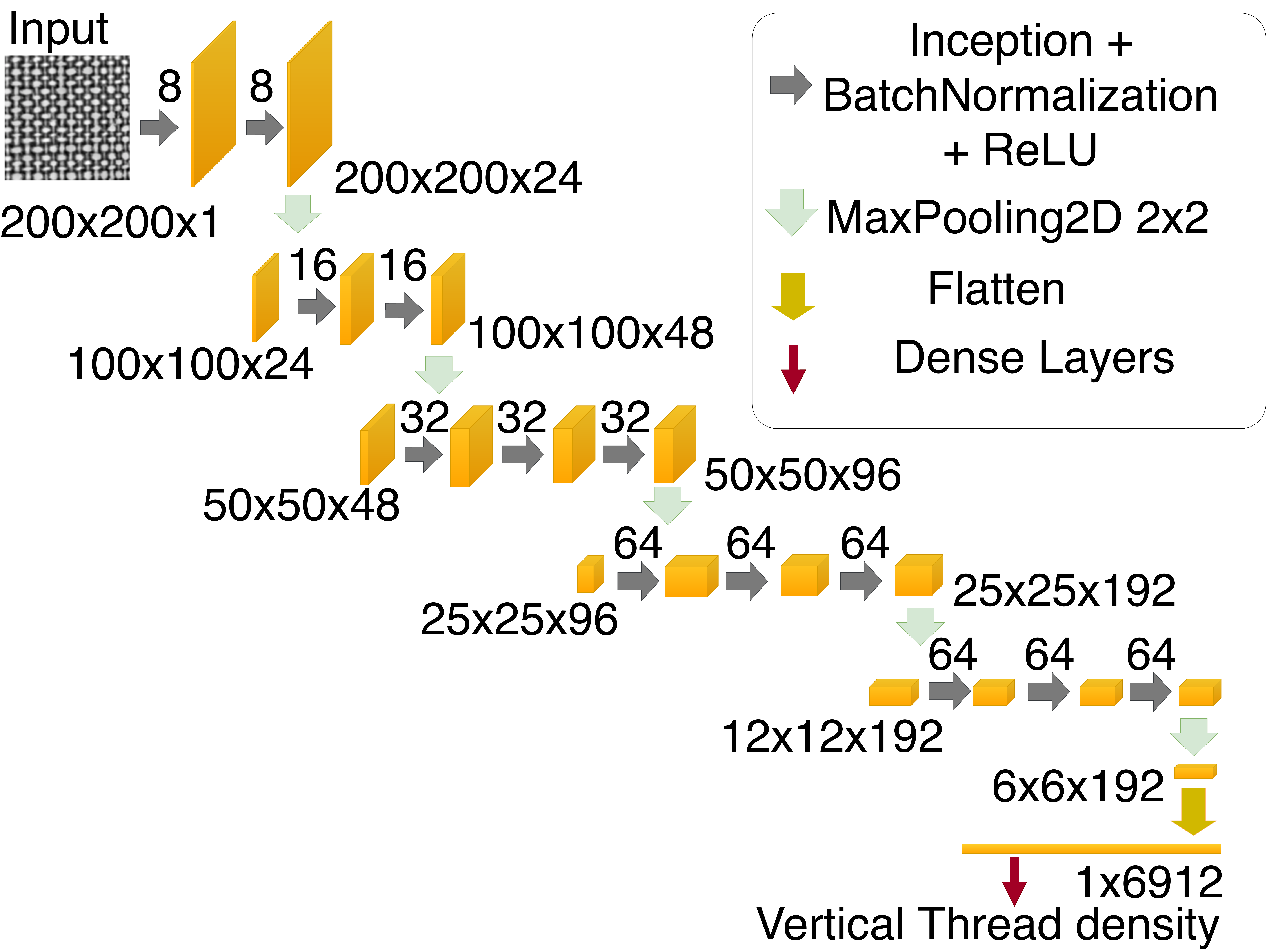}
\caption{Inception VGG-based regression model.} \label{fig:VGG_REG}
\end{figure}

\subsection{Inception Residual Regression Model}\label{ssec:Res}

\begin{figure}[htbp]
\centering
 \includegraphics[width=8cm]{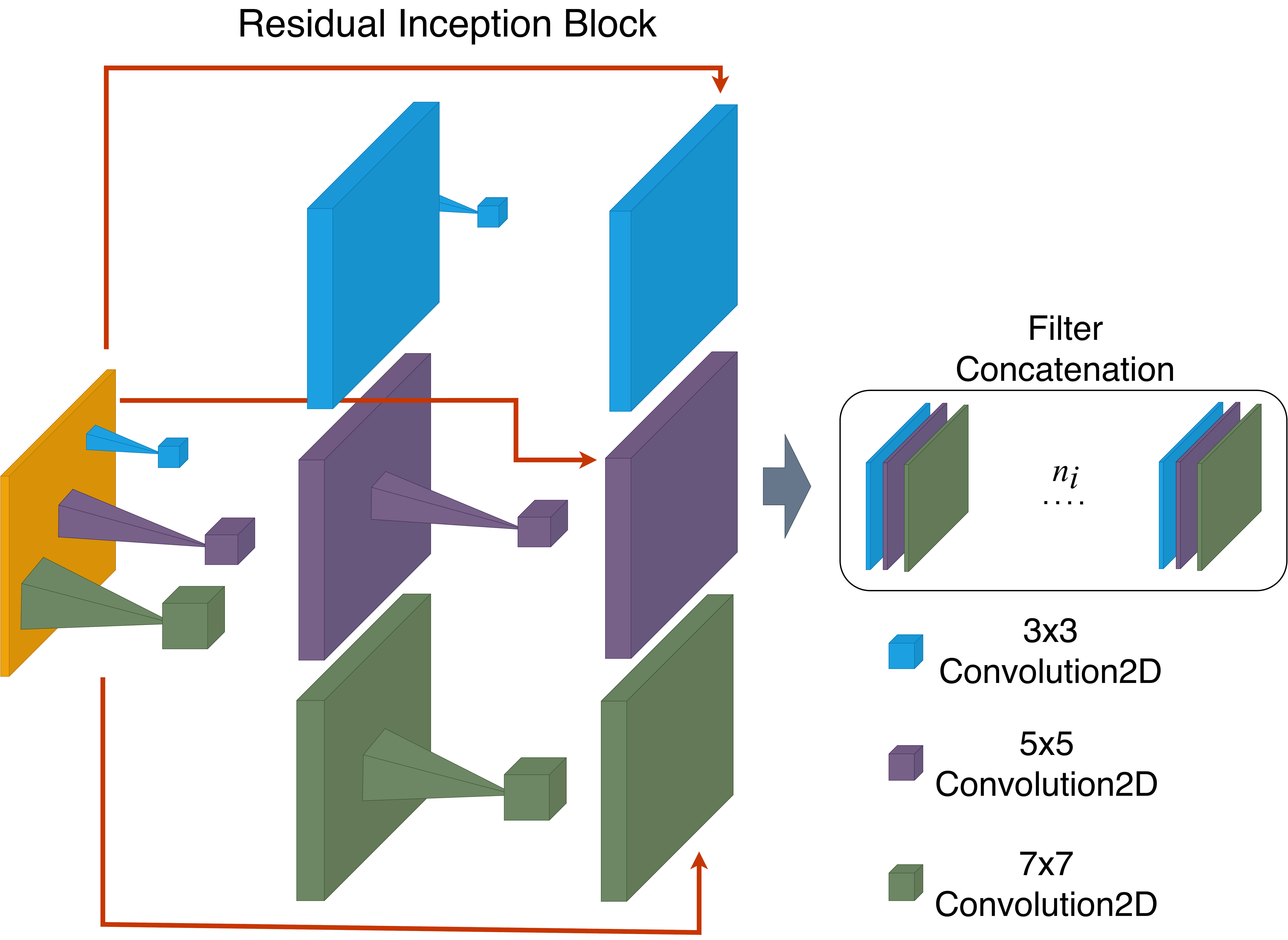}
\caption{Residual inception block based on the inception block in Fig. \ref{fig:Inception}.} \label{fig:ResBlock}
\end{figure}

\begin{figure}[htbp]
\centering
 \includegraphics[width=8.3cm]{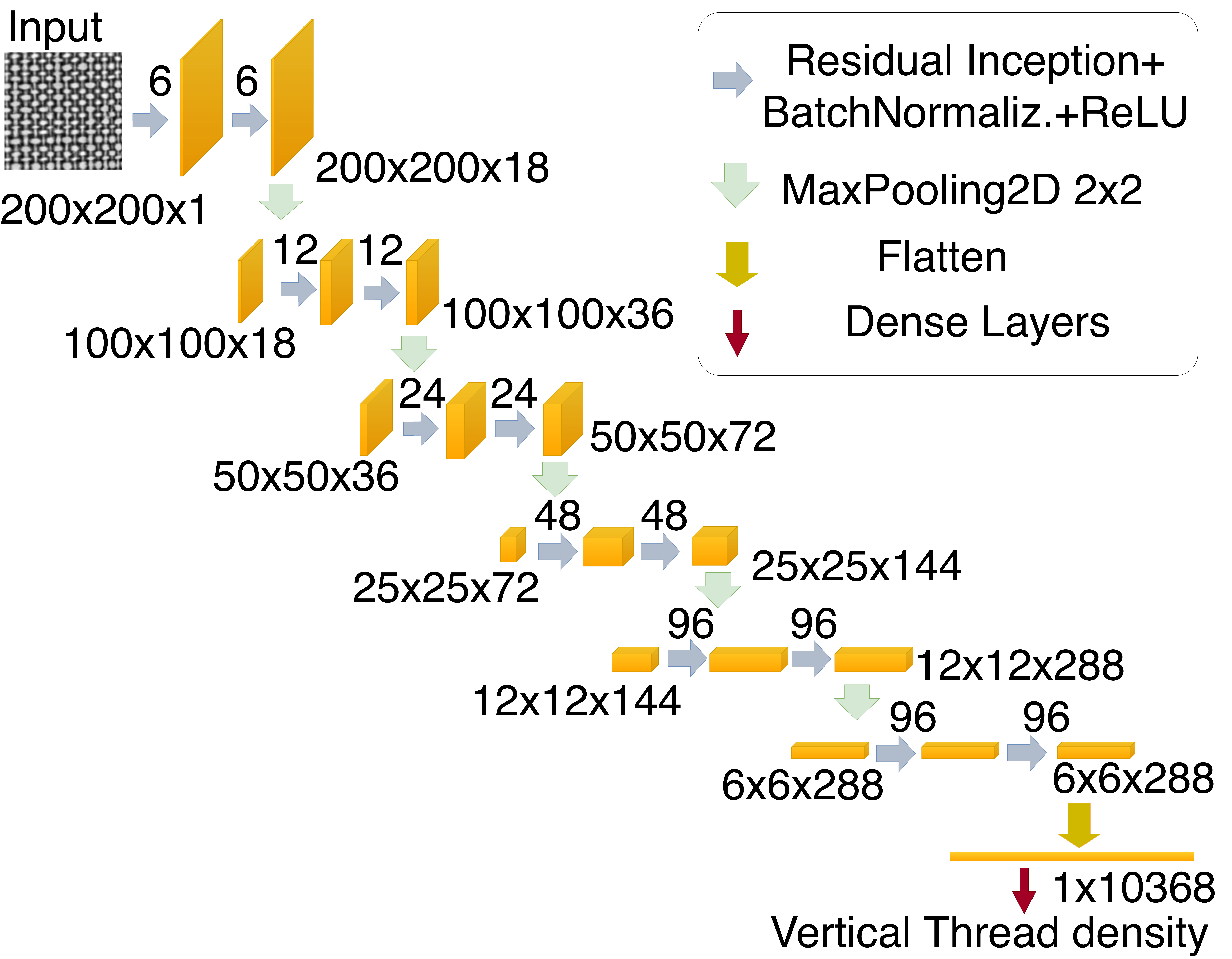}
\caption{Residual Inception Regression model.} \label{fig:RESREG}
\end{figure}

The last architecture developed incorporates the residual learning paradigm by modifying the inception block in the inception regression model in Section \ref{ssec:Reg} and Fig. \ref{fig:REG}. In this block we add the input to the output of the convolutions, see Fig. \ref{fig:ResBlock}. The full model has over 10 million parameters. In Fig. \ref{fig:RESREG} we include the scheme of the model.

\section{Training and Testing Results}\label{sec:train}

\subsection{Setup}
The models described above were programmed using Python 3.9, CUDA 11.2, and Keras-Tensorflow 2.5.0. The input grayscale image size was $200 \times 200 \times 1$ and the default batch size was 32. We used the Adam optimizer \cite{Kingma15} with a default learning rate value of $1\cdot10^{-3}$. Early stopping is used with a latency of 65 epochs and a limit of 450 epochs. The code was run on an NVIDIA RTX A6000 where any training typically takes around six hours, however, the training time varies across the different models as later reported. 

We trained every model 10 times. In every training, the weights of the models were randomly initialized. The datasets were also different throughout the 10 training, as in the DA different random rotations were applied. The same datasets were used in all models. The normalized mean absolute error (NMAE) was used as loss function, where the error was normalized by the value of the label, i.e., the thread density. In each training, we randomly initialized the weights of the model and randomly set the rotations in the DA to generate the training and validation datasets (see \cite{Bejarano2023} for further details).

We also train the Inc-Dice model in \cite{Bejarano2023} including the improvements in Section \ref{sec:DG}. The best model found, the one with the lowest validation loss among the 10 trainings, was used in the comparisons. Besides, this result was also used in the U-Net based regression model in Section \ref{ssec:RegUnet}.

%
%
%

\subsection{Training and Testing}
\begin{figure}[htp]
\centering
\begin{tikzpicture}

\definecolor{brown1926061}{RGB}{192,60,61}
\definecolor{darkslategray38}{RGB}{38,38,38}
\definecolor{darkslategray61}{RGB}{61,61,61}
\definecolor{lightgray204}{RGB}{204,204,204}
\definecolor{mediumpurple147113178}{RGB}{147,113,178}
\definecolor{peru22412844}{RGB}{224,128,44}
\definecolor{seagreen5814558}{RGB}{58,145,58}
\definecolor{steelblue49115161}{RGB}{49,115,161}

\begin{axis}[
axis line style={lightgray204},
tick align=outside,
x grid style={lightgray204},
xlabel=\textcolor{darkslategray38}{Models},
xmin=-0.5, xmax=4.5,
xtick style={color=darkslategray38},
xtick={0,1,2,3,4},
xticklabels={Reg-Unet,Reg,Reg-VGG,Residual,IncDice},
y grid style={lightgray204},
ylabel=\textcolor{darkslategray38}{Validation NMAE (\%)},
ymajorgrids,
ymin=0.824078426620843, ymax=2.86754917854291,
ytick style={color=darkslategray38}
]
\path [draw=darkslategray61, fill=steelblue49115161, semithick]
(axis cs:-0.4,2.2984584545917)
--(axis cs:0.4,2.2984584545917)
--(axis cs:0.4,2.53669090687491)
--(axis cs:-0.4,2.53669090687491)
--(axis cs:-0.4,2.2984584545917)
--cycle;
\path [draw=darkslategray61, fill=peru22412844, semithick]
(axis cs:0.6,0.947334674167854)
--(axis cs:1.4,0.947334674167854)
--(axis cs:1.4,0.980838016621642)
--(axis cs:0.6,0.980838016621642)
--(axis cs:0.6,0.947334674167854)
--cycle;
\path [draw=darkslategray61, fill=seagreen5814558, semithick]
(axis cs:1.6,0.953419044459612)
--(axis cs:2.4,0.953419044459612)
--(axis cs:2.4,1.02836175359939)
--(axis cs:1.6,1.02836175359939)
--(axis cs:1.6,0.953419044459612)
--cycle;
\path [draw=darkslategray61, fill=brown1926061, semithick]
(axis cs:2.6,1.00322149137223)
--(axis cs:3.4,1.00322149137223)
--(axis cs:3.4,1.03937849935059)
--(axis cs:2.6,1.03937849935059)
--(axis cs:2.6,1.00322149137223)
--cycle;
\path [draw=darkslategray61, fill=mediumpurple147113178, semithick]
(axis cs:3.6,1.07047335086434)
--(axis cs:4.4,1.07047335086434)
--(axis cs:4.4,1.13296523841858)
--(axis cs:3.6,1.13296523841858)
--(axis cs:3.6,1.07047335086434)
--cycle;
\addplot [semithick, darkslategray61]
table {%
0 2.2984584545917
0 2.1299203338543
};
\addplot [semithick, darkslategray61]
table {%
0 2.53669090687491
0 2.77466414436464
};
\addplot [semithick, darkslategray61]
table {%
-0.2 2.1299203338543
0.2 2.1299203338543
};
\addplot [semithick, darkslategray61]
table {%
-0.2 2.77466414436464
0.2 2.77466414436464
};
\addplot [semithick, darkslategray61]
table {%
1 0.947334674167854
1 0.929467934758842
};
\addplot [semithick, darkslategray61]
table {%
1 0.980838016621642
1 0.981920920492706
};
\addplot [semithick, darkslategray61]
table {%
0.8 0.929467934758842
1.2 0.929467934758842
};
\addplot [semithick, darkslategray61]
table {%
0.8 0.981920920492706
1.2 0.981920920492706
};
\addplot [black, mark=diamond*, mark size=3, mark options={solid,fill=darkslategray61}, only marks]
table {%
1 1.03497945645865
};
\addplot [semithick, darkslategray61]
table {%
2 0.953419044459612
2 0.916963460799119
};
\addplot [semithick, darkslategray61]
table {%
2 1.02836175359939
2 1.12179146130924
};
\addplot [semithick, darkslategray61]
table {%
1.8 0.916963460799119
2.2 0.916963460799119
};
\addplot [semithick, darkslategray61]
table {%
1.8 1.12179146130924
2.2 1.12179146130924
};
\addplot [semithick, darkslategray61]
table {%
3 1.00322149137223
3 0.953127496671489
};
\addplot [semithick, darkslategray61]
table {%
3 1.03937849935059
3 1.08444552072626
};
\addplot [semithick, darkslategray61]
table {%
2.8 0.953127496671489
3.2 0.953127496671489
};
\addplot [semithick, darkslategray61]
table {%
2.8 1.08444552072626
3.2 1.08444552072626
};
\addplot [semithick, darkslategray61]
table {%
4 1.07047335086434
4 1.03664159163596
};
\addplot [semithick, darkslategray61]
table {%
4 1.13296523841858
4 1.2191520301375
};
\addplot [semithick, darkslategray61]
table {%
3.8 1.03664159163596
4.2 1.03664159163596
};
\addplot [semithick, darkslategray61]
table {%
3.8 1.2191520301375
4.2 1.2191520301375
};
\addplot [black, mark=diamond*, mark size=3, mark options={solid,fill=darkslategray61}, only marks]
table {%
4 1.25900291720206
};
\addplot [semithick, darkslategray61]
table {%
-0.4 2.46143520423071
0.4 2.46143520423071
};
\addplot [semithick, darkslategray61]
table {%
0.6 0.97304395178237
1.4 0.97304395178237
};
\addplot [semithick, darkslategray61]
table {%
1.6 1.00145276360633
2.4 1.00145276360633
};
\addplot [semithick, darkslategray61]
table {%
2.6 1.01684802096654
3.4 1.01684802096654
};
\addplot [semithick, darkslategray61]
table {%
3.6 1.11961276712113
4.4 1.11961276712113
};
\end{axis}

\end{tikzpicture}
\caption{Box diagram of validation loss, i.e., the average NMAE (\%) of the density estimations for the validation set, for the 10 trainings and the 4 regression models considered plus the Inc-Dice segmentation DL. Outliers are represented with diamonds.}  
\label{fig:ValReg}
\end{figure}
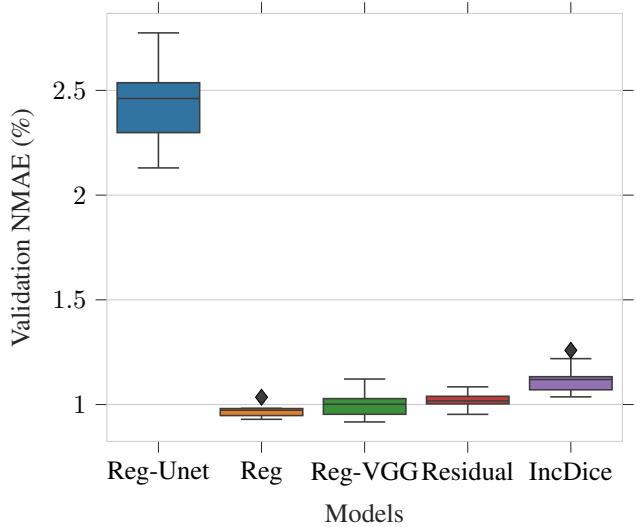

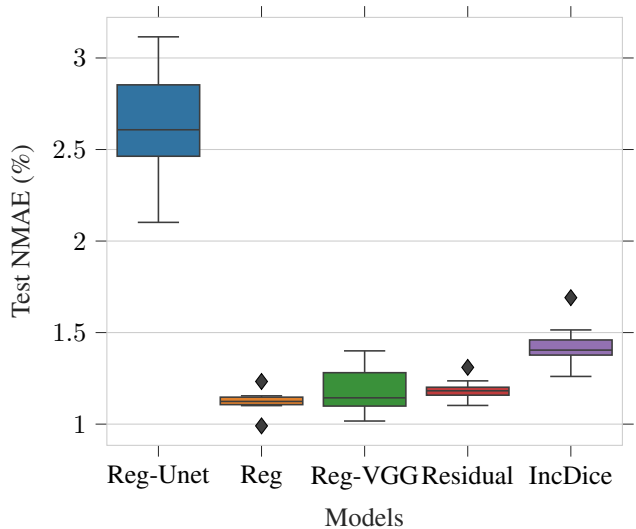
\begin{figure}[htbp]
\centering
\begin{tikzpicture}

\definecolor{brown1926061}{RGB}{192,60,61}
\definecolor{darkslategray38}{RGB}{38,38,38}
\definecolor{darkslategray61}{RGB}{61,61,61}
\definecolor{lightgray204}{RGB}{204,204,204}
\definecolor{mediumpurple147113178}{RGB}{147,113,178}
\definecolor{peru22412844}{RGB}{224,128,44}
\definecolor{seagreen5814558}{RGB}{58,145,58}
\definecolor{steelblue49115161}{RGB}{49,115,161}

\begin{axis}[
axis line style={lightgray204},
tick align=outside,
x grid style={lightgray204},
xlabel=\textcolor{darkslategray38}{Models},
xmin=-0.5, xmax=4.5,
xtick style={color=darkslategray38},
xtick={0,1,2,3,4},
xticklabels={Reg-Unet,Reg,Reg-VGG,Residual,IncDice},
y grid style={lightgray204},
ylabel=\textcolor{darkslategray38}{Test NMAE (\%)},
ymajorgrids,
ymin=0.884131081000669, ymax=3.22168287187362,
ytick style={color=darkslategray38}
]
\path [draw=darkslategray61, fill=steelblue49115161, semithick]
(axis cs:-0.4,2.46321942449928)
--(axis cs:0.4,2.46321942449928)
--(axis cs:0.4,2.85320356926887)
--(axis cs:-0.4,2.85320356926887)
--(axis cs:-0.4,2.46321942449928)
--cycle;
\path [draw=darkslategray61, fill=peru22412844, semithick]
(axis cs:0.6,1.10643239003629)
--(axis cs:1.4,1.10643239003629)
--(axis cs:1.4,1.14736930873591)
--(axis cs:0.6,1.14736930873591)
--(axis cs:0.6,1.10643239003629)
--cycle;
\path [draw=darkslategray61, fill=seagreen5814558, semithick]
(axis cs:1.6,1.09855208092833)
--(axis cs:2.4,1.09855208092833)
--(axis cs:2.4,1.28099441629875)
--(axis cs:1.6,1.28099441629875)
--(axis cs:1.6,1.09855208092833)
--cycle;
\path [draw=darkslategray61, fill=brown1926061, semithick]
(axis cs:2.6,1.15811176892021)
--(axis cs:3.4,1.15811176892021)
--(axis cs:3.4,1.20159203112214)
--(axis cs:2.6,1.20159203112214)
--(axis cs:2.6,1.15811176892021)
--cycle;
\path [draw=darkslategray61, fill=mediumpurple147113178, semithick]
(axis cs:3.6,1.37707843452498)
--(axis cs:4.4,1.37707843452498)
--(axis cs:4.4,1.45982205890896)
--(axis cs:3.6,1.45982205890896)
--(axis cs:3.6,1.37707843452498)
--cycle;
\addplot [semithick, darkslategray61]
table {%
0 2.46321942449928
0 2.10203285079398
};
\addplot [semithick, darkslategray61]
table {%
0 2.85320356926887
0 3.11543051774303
};
\addplot [semithick, darkslategray61]
table {%
-0.2 2.10203285079398
0.2 2.10203285079398
};
\addplot [semithick, darkslategray61]
table {%
-0.2 3.11543051774303
0.2 3.11543051774303
};
\addplot [semithick, darkslategray61]
table {%
1 1.10643239003629
1 1.10111523089775
};
\addplot [semithick, darkslategray61]
table {%
1 1.14736930873591
1 1.15455019371843
};
\addplot [semithick, darkslategray61]
table {%
0.8 1.10111523089775
1.2 1.10111523089775
};
\addplot [semithick, darkslategray61]
table {%
0.8 1.15455019371843
1.2 1.15455019371843
};
\addplot [black, mark=diamond*, mark size=3, mark options={solid,fill=darkslategray61}, only marks]
table {%
1 0.990383435131258
1 1.23289321681174
};
\addplot [semithick, darkslategray61]
table {%
2 1.09855208092833
2 1.01692288173621
};
\addplot [semithick, darkslategray61]
table {%
2 1.28099441629875
2 1.40016160616263
};
\addplot [semithick, darkslategray61]
table {%
1.8 1.01692288173621
2.2 1.01692288173621
};
\addplot [semithick, darkslategray61]
table {%
1.8 1.40016160616263
2.2 1.40016160616263
};
\addplot [semithick, darkslategray61]
table {%
3 1.15811176892021
3 1.10204894654453
};
\addplot [semithick, darkslategray61]
table {%
3 1.20159203112214
3 1.23630047399395
};
\addplot [semithick, darkslategray61]
table {%
2.8 1.10204894654453
3.2 1.10204894654453
};
\addplot [semithick, darkslategray61]
table {%
2.8 1.23630047399395
3.2 1.23630047399395
};
\addplot [black, mark=diamond*, mark size=3, mark options={solid,fill=darkslategray61}, only marks]
table {%
3 1.30994310375212
};
\addplot [semithick, darkslategray61]
table {%
4 1.37707843452498
4 1.26050060008103
};
\addplot [semithick, darkslategray61]
table {%
4 1.45982205890896
4 1.51452619650304
};
\addplot [semithick, darkslategray61]
table {%
3.8 1.26050060008103
4.2 1.26050060008103
};
\addplot [semithick, darkslategray61]
table {%
3.8 1.51452619650304
4.2 1.51452619650304
};
\addplot [black, mark=diamond*, mark size=3, mark options={solid,fill=darkslategray61}, only marks]
table {%
4 1.6908036979183
};
\addplot [semithick, darkslategray61]
table {%
-0.4 2.60785504194306
0.4 2.60785504194306
};
\addplot [semithick, darkslategray61]
table {%
0.6 1.12386538083148
1.4 1.12386538083148
};
\addplot [semithick, darkslategray61]
table {%
1.6 1.14365093431843
2.4 1.14365093431843
};
\addplot [semithick, darkslategray61]
table {%
2.6 1.1820777529085
3.4 1.1820777529085
};
\addplot [semithick, darkslategray61]
table {%
3.6 1.40433455532713
4.4 1.40433455532713
};
\end{axis}

\end{tikzpicture}
\caption{Box diagram of the average NMAE (\%) of the density estimations for the test set, for the 10 trainings and the 4 regression models considered and the Inc-Dice segmentation DL. Outliers are represented with diamonds.} \label{fig:TestError}
\end{figure}

In Fig. \ref{fig:ValReg} we include the box diagram with the results of the validation loss for the regression approaches developed: the regression based on the encoder of the U-Net in Section \ref{ssec:RegUnet}, inception regression in Section \ref{ssec:RegUnet}, the VGG based regression model in Section \ref{ssec:ResVGG} and the residual version of the inception regression in Section \ref{ssec:Res}, denoted by Reg-Unet, Reg, Reg-VGG, and Reg-Res, respectively. In Fig. \ref{fig:TestError} we include the results for the same models but for the test data. 

In view of the validation NMAE values in Fig. \ref{fig:ValReg} it can be concluded that the Reg-Unet approach is not a good option, i.e., reusing the encoder of the segmentation is not useful. On the contrary, the Reg, Reg-VGG, and Reg-Res approaches provide good low values with low variation of the results for the 10 trainings. The Inc-Dice model in \cite{Bejarano2023} with lowest NMAE achieved 1.12\% while with the improvements in Section \ref{sec:DG} we have a NMAE of 1.04\%, i.e., we have a 0.8\% improvement. The Inc-Dice model, however, presents an important increase when evaluated with test data, see Fig. \ref{fig:TestError}. 

Among the evaluated methods the Reg-VGG exhibits the lowest value of NMAE for the validation set, 0.92\%. We will use this model in the analysis of the canvases later in this work. Hence, compared to the segmentation DL approach the regression DL method has a 0.12\% lower error and does not need further signal processing as it directly computes the thread densities. 

When evaluated in the test set, see Fig. \ref{fig:TestError},  regression DL models have a very much better error value than the Inc-Dice.  Specifically, the NMAE in test has decreased from 1.51\% of the Inc-Dice to 1.02\% of the Reg-VGG for the best model in validation. The NMAE of the Inc-Dice as in \cite{Bejarano2023}, i.e., without the improvements in the generation of the dataset, increases to 1.61\%. %
 
%
%
%
%
%

We end up providing the average time needed in the training of the models, included in Tab. \ref{tab:comptimes}. We observe that the Reg-Unet model needs half the time the Reg or the Reg-VGG models, while the Reg-Res needs approximately double the time.
\begin{table}[htb]
\caption{Average computation times, in hours, of the training of the regression DL models.}\label{tab:comptimes}
\begin{tabular}{c | c c c c c}
\toprule
    Method             & Reg-Unet& Reg & Reg-VGG & Reg-Res  \\
\midrule
Time (hours) & 1.71 & 3.73 & 3.91& 6.17\\
\bottomrule
\end{tabular}
\end{table}

\subsection{Ablation Study of Preprocessing}\label{ssec:Ablation}

To analyze the impact of improvements in Section \ref{sec:DG} we perform an ablation study where for the best model found, the Reg-VGG, we train and report the value of the NMAE for the test set when each of the improvements is not used. We include the NMAE for the inception VGG regression and this same model with no equalization, with kernels of fixed size, when central images are not exploited and when maximum random rotation angles in the DA are doubled, denoted by Reg, No-Eq, Fix-Win, No-Crop, and 2xAngle, respectively.

\begin{figure}[htp]
\centering
\begin{tikzpicture}

\definecolor{brown1926061}{RGB}{192,60,61}
\definecolor{darkslategray38}{RGB}{38,38,38}
\definecolor{darkslategray61}{RGB}{61,61,61}
\definecolor{lightgray204}{RGB}{204,204,204}
\definecolor{mediumpurple147113178}{RGB}{147,113,178}
\definecolor{peru22412844}{RGB}{224,128,44}
\definecolor{seagreen5814558}{RGB}{58,145,58}
\definecolor{steelblue49115161}{RGB}{49,115,161}

\begin{axis}[
axis line style={lightgray204},
tick align=outside,
x grid style={lightgray204},
xlabel=\textcolor{darkslategray38}{Models},
xmin=-0.5, xmax=4.5,
xtick style={color=darkslategray38},
xtick={0,1,2,3,4},
xticklabels={Reg,No-Eq,Fix-Win,No-Crop,2xAngle},
y grid style={lightgray204},
ylabel=\textcolor{darkslategray38}{Test NMAE (\%)},
ymajorgrids,
ymin=0.984116965331971, ymax=1.70584712622529,
ytick style={color=darkslategray38}
]
\path [draw=darkslategray61, fill=steelblue49115161, semithick]
(axis cs:-0.4,1.09855208092833)
--(axis cs:0.4,1.09855208092833)
--(axis cs:0.4,1.28099441629875)
--(axis cs:-0.4,1.28099441629875)
--(axis cs:-0.4,1.09855208092833)
--cycle;
\path [draw=darkslategray61, fill=peru22412844, semithick]
(axis cs:0.6,1.11949588552307)
--(axis cs:1.4,1.11949588552307)
--(axis cs:1.4,1.2138485491507)
--(axis cs:0.6,1.2138485491507)
--(axis cs:0.6,1.11949588552307)
--cycle;
\path [draw=darkslategray61, fill=seagreen5814558, semithick]
(axis cs:1.6,1.17499377156956)
--(axis cs:2.4,1.17499377156956)
--(axis cs:2.4,1.30582618907354)
--(axis cs:1.6,1.30582618907354)
--(axis cs:1.6,1.17499377156956)
--cycle;
\path [draw=darkslategray61, fill=brown1926061, semithick]
(axis cs:2.6,1.09700971152536)
--(axis cs:3.4,1.09700971152536)
--(axis cs:3.4,1.14710762198309)
--(axis cs:2.6,1.14710762198309)
--(axis cs:2.6,1.09700971152536)
--cycle;
\path [draw=darkslategray61, fill=mediumpurple147113178, semithick]
(axis cs:3.6,1.08039564137693)
--(axis cs:4.4,1.08039564137693)
--(axis cs:4.4,1.20129640303458)
--(axis cs:3.6,1.20129640303458)
--(axis cs:3.6,1.08039564137693)
--cycle;
\addplot [semithick, darkslategray61]
table {%
0 1.09855208092833
0 1.01692288173621
};
\addplot [semithick, darkslategray61]
table {%
0 1.28099441629875
0 1.40062535347857
};
\addplot [semithick, darkslategray61]
table {%
-0.2 1.01692288173621
0.2 1.01692288173621
};
\addplot [semithick, darkslategray61]
table {%
-0.2 1.40062535347857
0.2 1.40062535347857
};
\addplot [semithick, darkslategray61]
table {%
1 1.11949588552307
1 1.08517077122815
};
\addplot [semithick, darkslategray61]
table {%
1 1.2138485491507
1 1.22047296813172
};
\addplot [semithick, darkslategray61]
table {%
0.8 1.08517077122815
1.2 1.08517077122815
};
\addplot [semithick, darkslategray61]
table {%
0.8 1.22047296813172
1.2 1.22047296813172
};
\addplot [black, mark=diamond*, mark size=3, mark options={solid,fill=darkslategray61}, only marks]
table {%
1 1.61065048266513
};
\addplot [semithick, darkslategray61]
table {%
2 1.17499377156956
2 1.08959106212327
};
\addplot [semithick, darkslategray61]
table {%
2 1.30582618907354
2 1.40531247708067
};
\addplot [semithick, darkslategray61]
table {%
1.8 1.08959106212327
2.2 1.08959106212327
};
\addplot [semithick, darkslategray61]
table {%
1.8 1.40531247708067
2.2 1.40531247708067
};
\addplot [black, mark=diamond*, mark size=3, mark options={solid,fill=darkslategray61}, only marks]
table {%
2 1.67304120982105
};
\addplot [semithick, darkslategray61]
table {%
3 1.09700971152536
3 1.04021742731695
};
\addplot [semithick, darkslategray61]
table {%
3 1.14710762198309
3 1.20111650271348
};
\addplot [semithick, darkslategray61]
table {%
2.8 1.04021742731695
3.2 1.04021742731695
};
\addplot [semithick, darkslategray61]
table {%
2.8 1.20111650271348
3.2 1.20111650271348
};
\addplot [semithick, darkslategray61]
table {%
4 1.08039564137693
4 1.06177310400461
};
\addplot [semithick, darkslategray61]
table {%
4 1.20129640303458
4 1.26736836015092
};
\addplot [semithick, darkslategray61]
table {%
3.8 1.06177310400461
4.2 1.06177310400461
};
\addplot [semithick, darkslategray61]
table {%
3.8 1.26736836015092
4.2 1.26736836015092
};
\addplot [semithick, darkslategray61]
table {%
-0.4 1.14381436197598
0.4 1.14381436197598
};
\addplot [semithick, darkslategray61]
table {%
0.6 1.19247745757901
1.4 1.19247745757901
};
\addplot [semithick, darkslategray61]
table {%
1.6 1.24191562182427
2.4 1.24191562182427
};
\addplot [semithick, darkslategray61]
table {%
2.6 1.13457615202707
3.4 1.13457615202707
};
\addplot [semithick, darkslategray61]
table {%
3.6 1.09212248244153
4.4 1.09212248244153
};
\end{axis}

\end{tikzpicture}
\caption{Box diagram of the average NMAE (\%) of the density estimations for the test dataset in the ablation study: the regression model considered (Reg) without equalization (No-Eq), a pre-processing with fixed window (Fix-Win), removing images taken from the central parts of the labeled samples (No-Crop), and limiting the maximum value of the random angle in DA (2xAngle). In each case, 10 different trainings were run. Outliers are represented with diamonds. }  
\label{fig:AblationRegTest}
\end{figure}
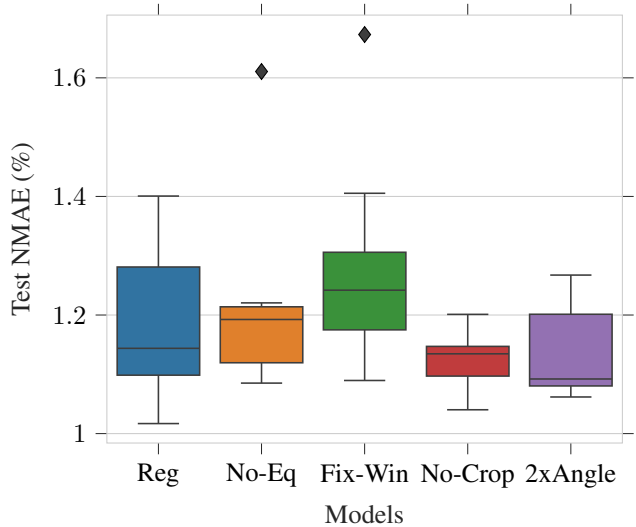

The equalization is quite relevant for the regression DL approach. In \cite{Bejarano2023} filtering to enhance the mean and variance of the input image was applied. Besides, large and low values were clipped. In the view of Fig. \ref{fig:AblationRegTest}, it can be concluded that the equalization improves the estimation of the thread densities. In practice, we found that the improvement is noticeable for some X-ray plates while is irrelevant for others. 
On the other hand, we slightly improve the NMAE using a variable window. Note that we will have larger reductions in the error for fabrics with optimal window sizes different from the fixed value used in \cite{Bejarano2023}. Put in other words, the variable window size is especially interesting for canvases with extreme values of threads density values. 
Regarding DA, adding central patches is also of interest as we are enlarging the data set. Besides, limiting the maximum angle of the random rotations in DA helps to achieve a solution with lower error. 
Finally, note that the minimum value of NMAE is provided by the inception VGG regression model with all improvements used and this is the model and weights selected to estimate densities in the following.


\section{Semi-Supervised Training}\label{sec:SS}


To improve the estimation in these scenarios, we propose a semi-supervised (SS) approach, see Alg. \ref{alg:SS}. The key of this SS method is to create, given a canvas to be analyzed, a dataset of patches where the FT and the regression DL provide similar estimates. Then, train the regression DL solution for some epochs with this dataset. This can be also viewed as a transfer learning, see \cite{Aradillas21} and references therein, where the starting point is the already trained inception VGG regression DL and we re-train it with this new dataset generated for the processed canvas. The main advantage is that we do not need to label any of the new images, as we use those where the FT and initial regression DL agree. After the SS algorithm, the refined model at the output is used to estimate the thread densities. Note that with this SS approach, we expect to improve the solution there were the FT provide good enough solutions and no improvement at all otherwise.

\begin{algorithm}
\caption{Semi-Supervised Learning}\label{alg:SS}
\textbf{Input:} Pre-processed patches and counts (predictions), trained regression DL model.
\begin{algorithmic}[1]
\State  Create a dataset with input images where the error in the estimation of FT and regression DL in both vertical and horizontal threads is under 4\%. 
%
%
\State 
To keep complexity bounded set a limit for the size, $N$, by randomly sampling instances in the created dataset.
\State Randomly split the dataset into 70\% and 30\% subsets, for training and validation, respectively.
\State Train the input model, with given weights as initial values, using the new dataset.  
\end{algorithmic}
\textbf{Output:} The input model with the values of the new weights.
\end{algorithm}

The proposed SS algorithm was applied row-wise as follows. The trained Reg-VGG and the FT were applied to the set of $F$ rows to obtain up to $N=60,000$ samples. These images were used to train the Reg-VGG initialized to the weights obtained in Sec. \ref{sec:train} with a learning rate set to $10^{-3}$, early stopping with 3 epochs of patience for a maximum of 20 epochs, and the NMAE as loss function. The last three layers in the dense network were frozen. 
We applied this SS regression DL approach to the four canvases in the validation set to further reduce the NMAE, from 0.92\% to 0.90\%. In test, the reduction is from 1.01\% to 0.94\%.

\begin{figure*}[htp]
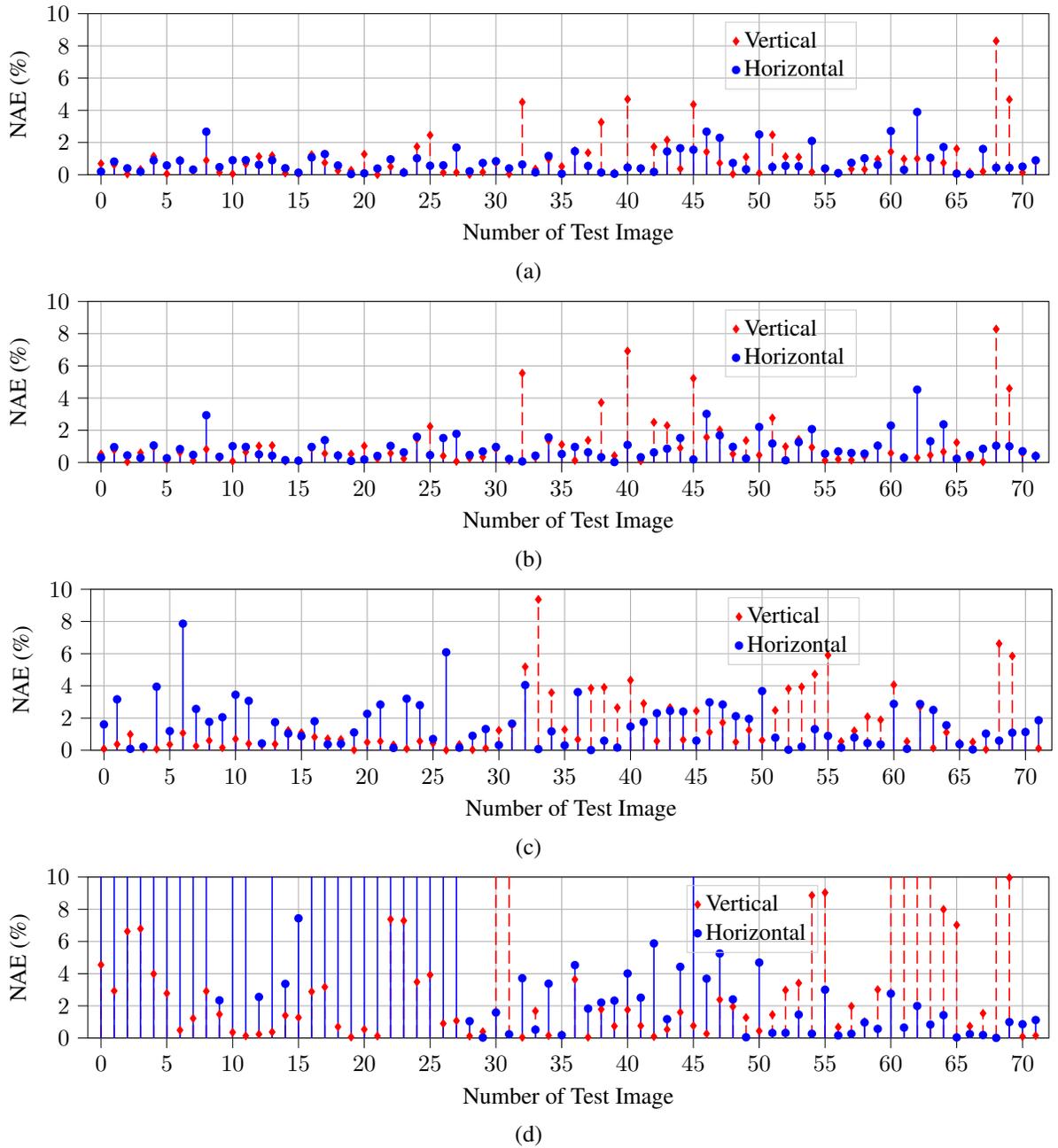

\centering
\begin{tabular}{c}
\input{ResultsTest_SS.tex} \\
 (a) \\
\input{RegressionVGGTest.tex} \\
 (b) \\
\input{model11BG4.tex}\\
 (c) \\ 
 \input{FT_TestFig.tex}
 \\
 (d) 
\end{tabular}
\caption{Vertical and horizontal thread density normalized absolute error (NAE) (\%) of every sample in the test dataset for (a) the best Reg model, (b) the Inc-Dice in \cite{Bejarano2023} and (c) the FT.} \label{fig:TestSamples}
\end{figure*}

For the sake of completeness, in  Fig. \ref{fig:TestSamples}.a we include the result of the normalized absolute error (NAE) in the vertical and horizontal density estimation for each test image and the best Reg model, i.e., the Reg-VGG with the lowest validation error, when SS is applied. In Fig. \ref{fig:TestSamples}.b we include the errors for the regression DL with no SS. In Fig. \ref{fig:TestSamples}.c and Fig. \ref{fig:TestSamples}.d we depict the error for the Inc-Dice as in \cite{Bejarano2023}  and the FT, respectively. The error of the FT is quite large. The first 28 images correspond to the same canvas and the method is unable to provide a valid estimation. Compared to the segmentation DL, the error of the regression DL decreases for most of the test samples. When using SS the errors are further reduced. Only two test samples exhibit an error greater than 5\% using the new regression approach. These errors could correspond to samples that were difficult to label, as threads were not easily observed.

\section{Canvas Analysis}\label{sec:exp}

In the following, we study a series of canvases. The X-ray plate image is first re-scaled to 200 pixels per cm, to get an image of size $r \times s$, the height by the width of the image in pixels, respectively. Then it is pre-processed. Patches of $1\times 1$ cm all over the canvas are the input to the regression DL. Depending on the spatial resolution needed for the threads maps, the patches can be overlapped from one patch to the next one by some pixels, $o$, either vertically or horizontally. Therefore, we will process $p\times q$ patches, where $p= \lfloor r\cdot (1-o)/200 \rfloor$ and $q=\lfloor s\cdot (1-o)/200 \rfloor$. In the SS algorithm, we set $F=40\cdot 200/(1-o)$ pixels, i.e., we process blocks of $40\times q$ patches. After processing the image we get two matrices with the estimated values for the vertical and horizontal thread densities. The matrices can be interpreted as images where the pixel value indexes a color, hereafter denoted by threads density maps. 
We compare threads density maps obtained with the new regression DL approach to the ones of other previous approaches as frequency analysis \cite{Simois18}, ATCA approach \cite{Maaten15} or segmentation DL \cite{Bejarano2023}. The color of any pixel of the color densities maps encodes the value of the thread density in thr/cm, estimated for the $1\times 1$ cm image area in that location, except for the  ATCA \cite{Maaten15} that computes densities from crossing points at that position.

\subsection{Semi-Supervised analysis with Ixion by Ribera}
\begin{figure*}[htb]
\centering
\begin{tabular}{cc}
\includegraphics[width=7cm]{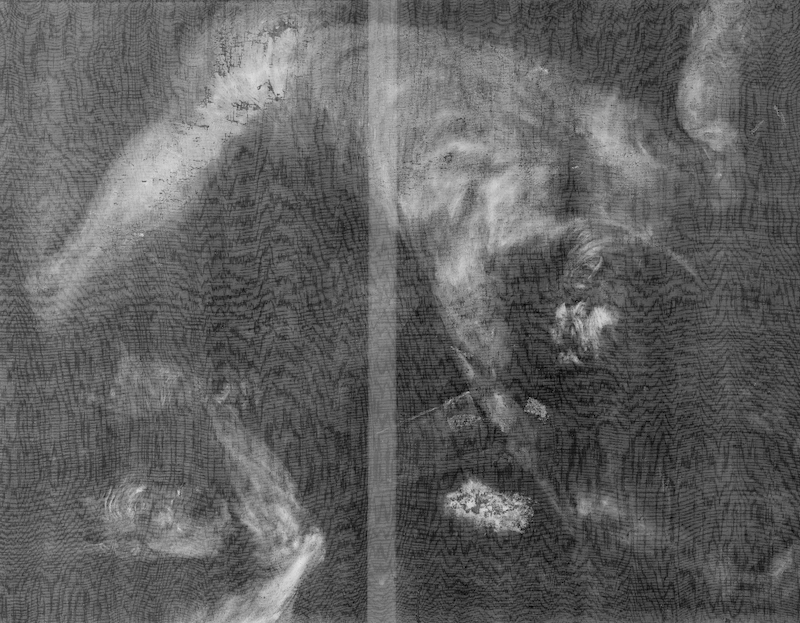} &  \includegraphics[width=7cm]{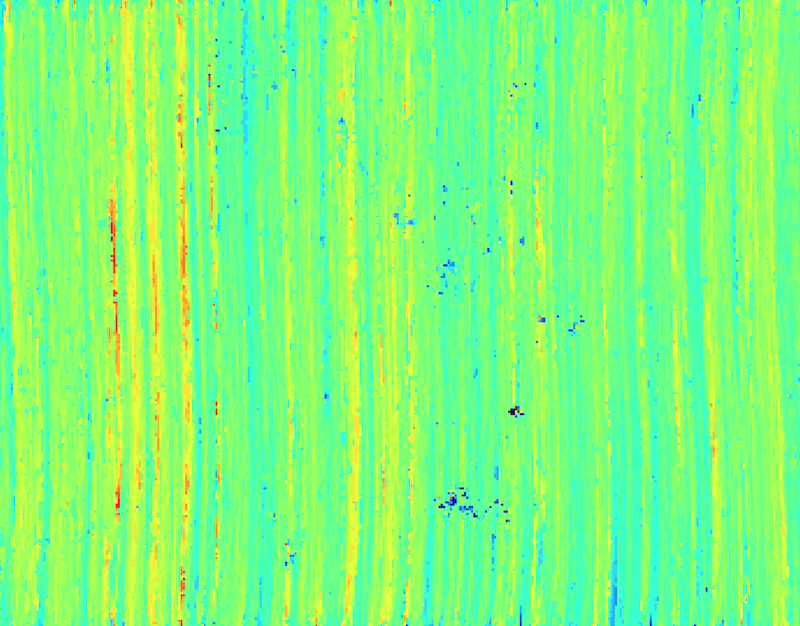}\\
 (a)& (b) \\
 \includegraphics[width=7cm]{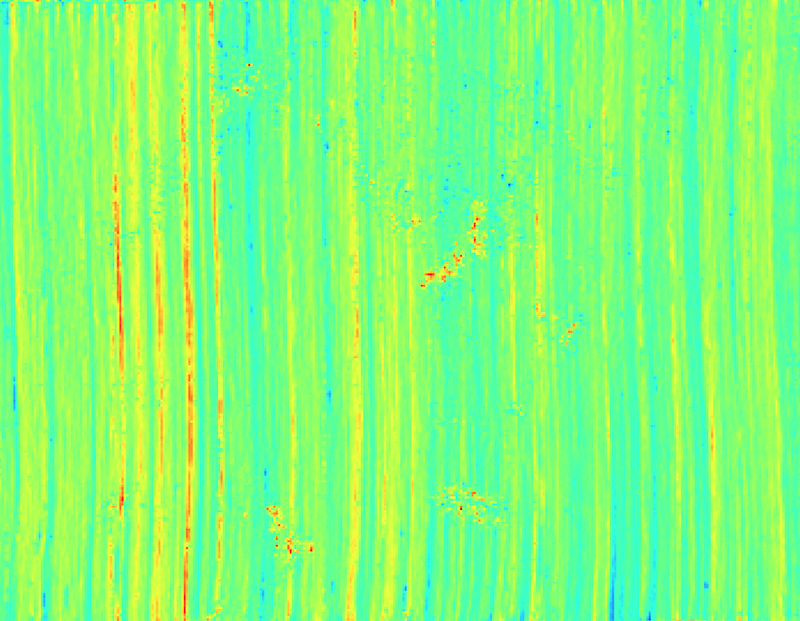}&\includegraphics[width=7cm]{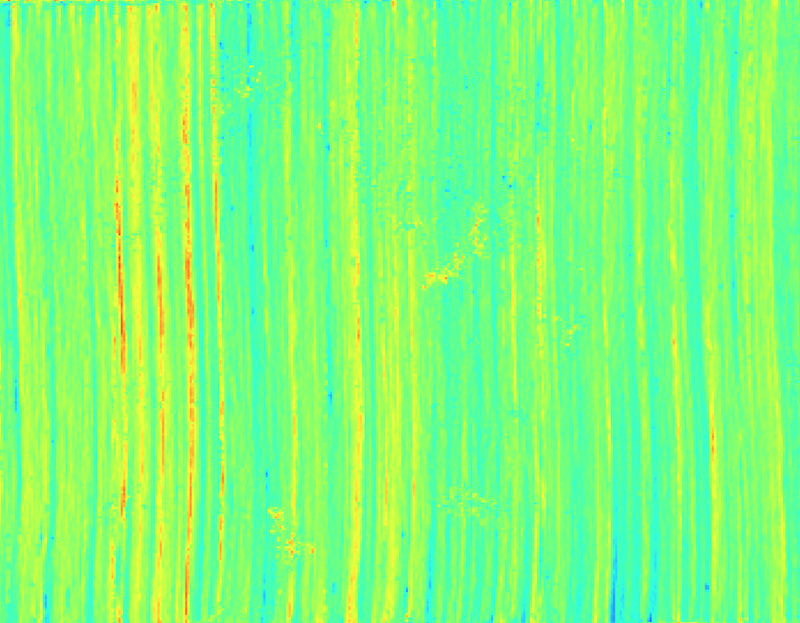}\\
 (c)& (d) \\
\end{tabular}
\caption{For the X-ray plate of Ixion by Ribera in (a), the color maps of vertical thread densities in thr/cm with (b) frequency analysis, (c) using regression DL and (d) using SS regression DL. Red-yellow colored areas are low densities ones while green-blue have larger densities.} \label{fig:Ribera}
\end{figure*}

The most used method for frequency analysis is FT, where no labeling is needed. If the fabric presents quite a uniform pattern where both weft and warp are clearly observed, the FT provides a quite good estimation of the thread densities. The painting Ixion by Ribera \cite{P001114} at the Museo N. del Prado is a large canvas, 3 m wide, with a very low threads density, around 6 thr/cm, and with a very uniform weave pattern. The result of the FT is included in Fig. \ref{fig:Ribera}.b. In Fig. \ref{fig:Ribera}.c the result of the inception regression can be observed, where patches are 50\% overlapped. The regression exhibits larger errors in saturated areas where the contrast is quite reduced. This is the case of the areas of the hip or shoulder, see Fig. \ref{fig:Ribera}.a. By using SS we improve the outcome in these areas, see Fig. \ref{fig:Ribera}.c.,  as the weights of the regression DL are refined with the FT result. 

\subsection{Comparison to ATCA for Poussin Paintings}
In this subsection we face the analysis of two canvases by Nicolas Poussin at The National Gallery, The Triumph of Pan \cite{NG6477} and The Triumph of Silenus \cite{NG42}. Radiographs of Triumph of Pan were scanned at 600 dpi and the ones from Triumph of Silenus were scanned at 1200 dpi. We stitched them into whole-painting images. After the analysis with the ATCA approach in \cite{Maaten15} they are known to come from the same bolt.

\begin{figure*}[tp]
\centering
\begin{tabular}{cccc}
\includegraphics[width=13cm]{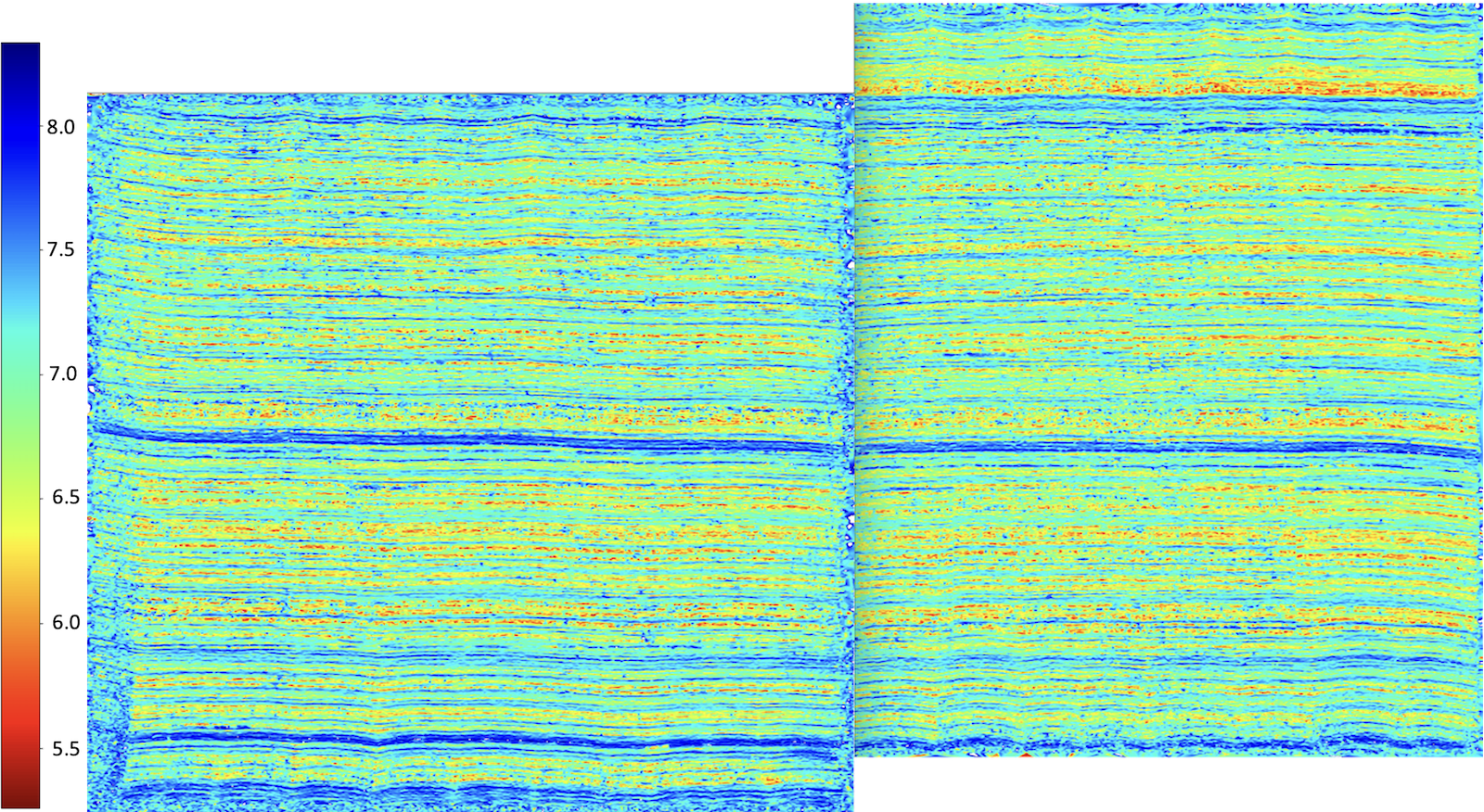} \\
 (a) \\
 \includegraphics[width=13cm]{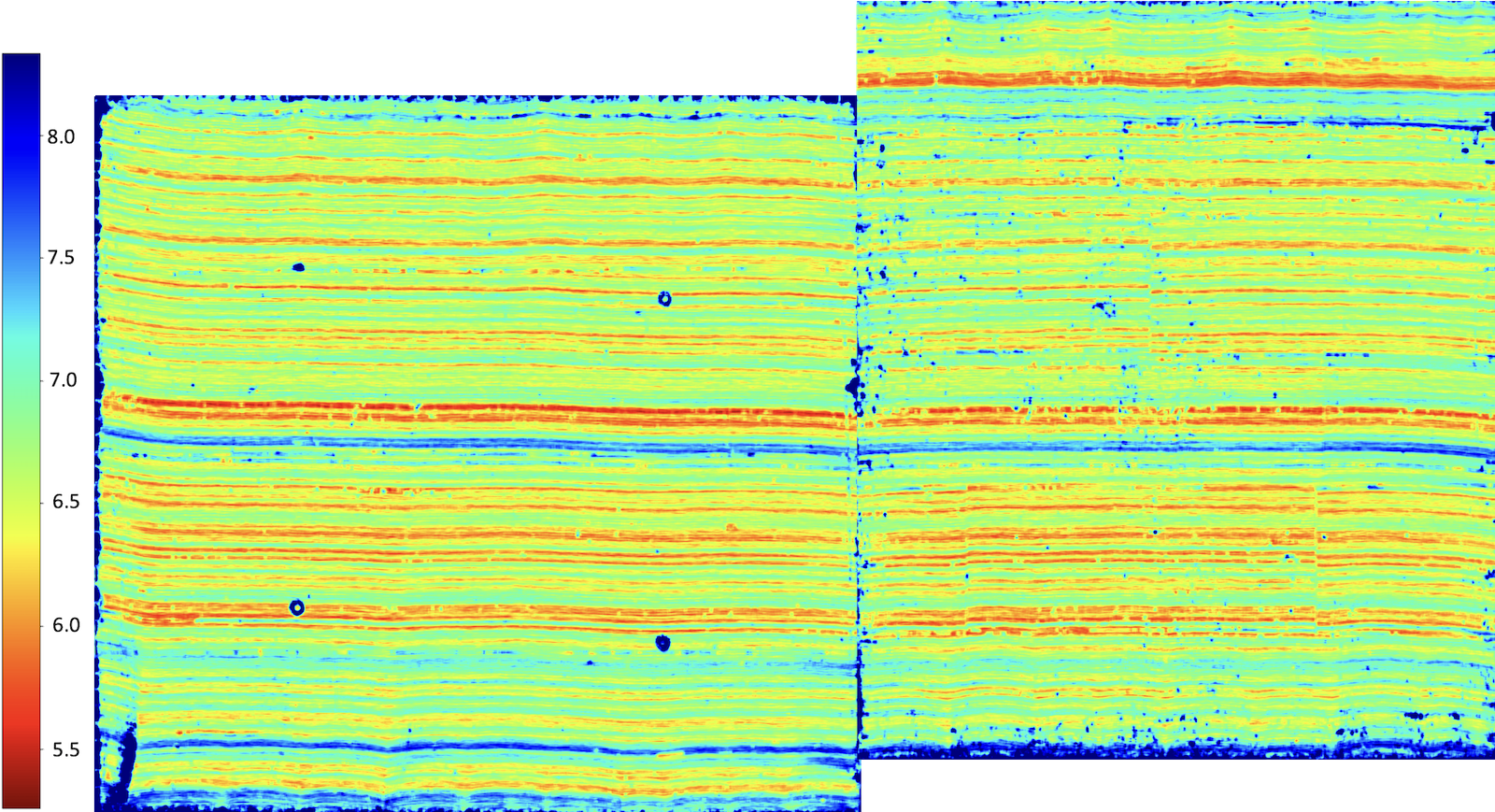}\\
 (b) \\
\end{tabular}
\caption{Match of horizontal thread densities, in thr/cm, for The Triumph of Pan, to the left, and The Triumph of Silenus, flipped vertically, to the right, using (a) ATCA in \cite{Maaten15} and (b) SS regression DL. Colormaps have been adjusted to the minimum and maximum values considered in \cite{Maaten15}.} \label{fig:Poussin}
\end{figure*}

In \cite{Maaten15} the authors labeled 11,954 thread crossings. For these images, they performed a grid search for the logistic regression regularization parameter, $\sigma^2$, and other parameters such as window size for filtering and thresholds were also set in the cleansing stage. By using the code in \cite{Maaten15b}, we performed the matching between them for the horizontal density estimations, included in Fig. \ref{fig:Poussin}.a. In Fig. \ref{fig:Poussin}.b we include the matching by using the novel proposed approach, the inception regression DL, with a 90\% overlap. In view of these results, we highlight the following.

\begin{figure}[!tp]
\centering
  \includegraphics[width=4.5cm]{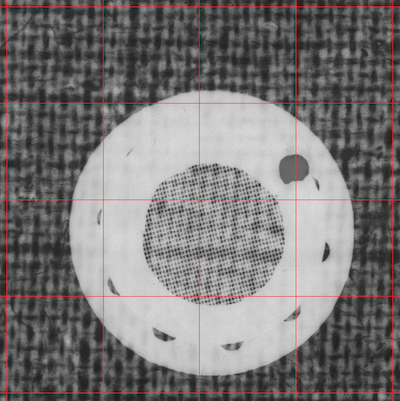}
\caption{Opaque object in the X-ray plate of the Triumph of Pan, by N. Poussin. The red grid included as a reference has squares of 1 cm sides.} \label{fig:circ}
\end{figure}

First, the ATCA provides a finer spatial resolution, as it is focused on the local distances between crossing points. Our approach is based on the estimation of the mean of the distances between crossing points within a 1 cm side square, providing a coarser estimation. 
Second, it can be observed that for higher densities the ATCA method better estimates the distances, and for lower values, the outcome is quite noisy. Hence, matching the paintings in areas with lower thread counting is much harder. The proposed approach provides good results in the whole range.
Third, while in the ATCA \textit{thousands of crossing points were needed to be labeled} by using the regression SS DL model no extra manual labeling is needed.
Finally, in Fig. \ref{fig:Poussin}.b and for the Triumph of Pan, to the left, there are four circle-shaped artifacts (in blue). In the original X-ray plate, we found some quite opaque objects, see in Fig \ref{fig:circ} for a detail of one of them in the upper right, where red lines, separated by 1 cm, are included as a reference.  

\subsection{Comparison to Inc-Dice of portraits by Velazquez}
\begin{figure}[!tp]
\centering
\begin{tabular}{cccc}
 \includegraphics[width=3.57cm]{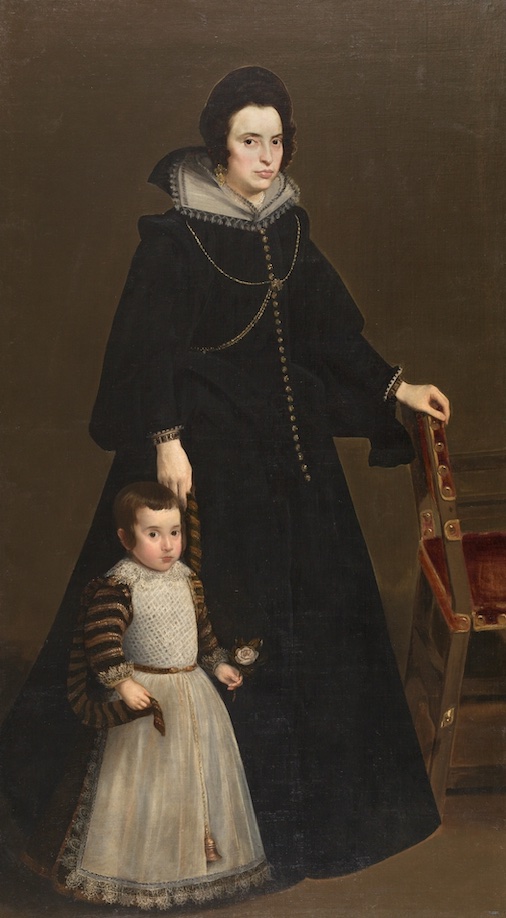}& 
  \includegraphics[width=3.61cm]{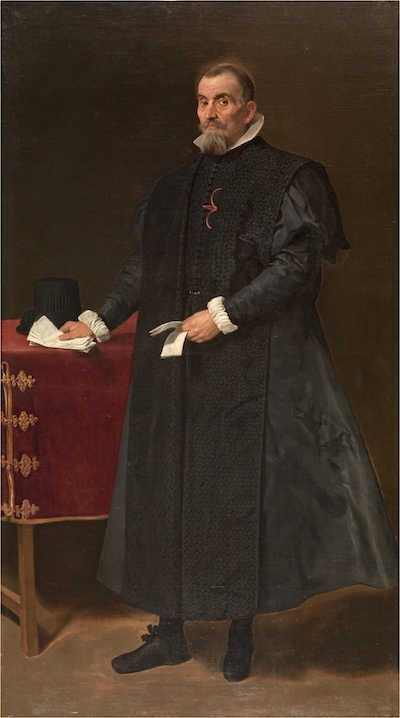}\\
(a) & (b) 
\end{tabular}
\caption{Paintings by Velázquez (a) Antonia de Ipeñarrieta y Galdós and her Son, Luis \cite{P001196} and (b) Diego del Corral y Arellano \cite{P001195}.} \label{fig:P0119X}
\end{figure}

We next analyze two canvases by Velázquez. On the one hand, Antonia de Ipeñarrieta y Galdós and Her Son Luis (P001196) \cite{P001196} and, on the other, Diego del Corral y Arellano (P001195) \cite{P001195}, at the Museo N. del Prado, in Fig. \ref{fig:P0119X}. In this couple of canvases, husband and wife were portrayed and it is conjectured that both were painted at the same time on fabrics from the same roll. Both plates have a medium-high density of threads and this type of fabric is actually problematic when frequency analyses are used. The segmentation DL approach already exhibited good performance and a correspondence between the horizontal density maps of both paintings was found, see \cite{Bejarano2023}. Therefore, both fabrics came from the same bolt. 

We compare the results of the regression and segmentation DL approaches.  Patches from Velazquez plates are overlapped 75\% horizontally and vertically. Therefore, we expect to have 16 pixels in the resulting density maps for every cm$^2$ in the fabric.

\begin{figure}[!tp]
\centering
\begin{tabular}{cccc}
 \includegraphics[width=3.235cm]{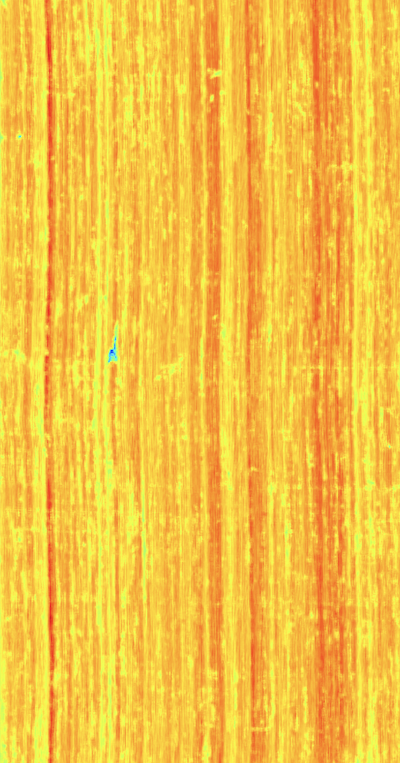}&
  \includegraphics[width=4.1cm]{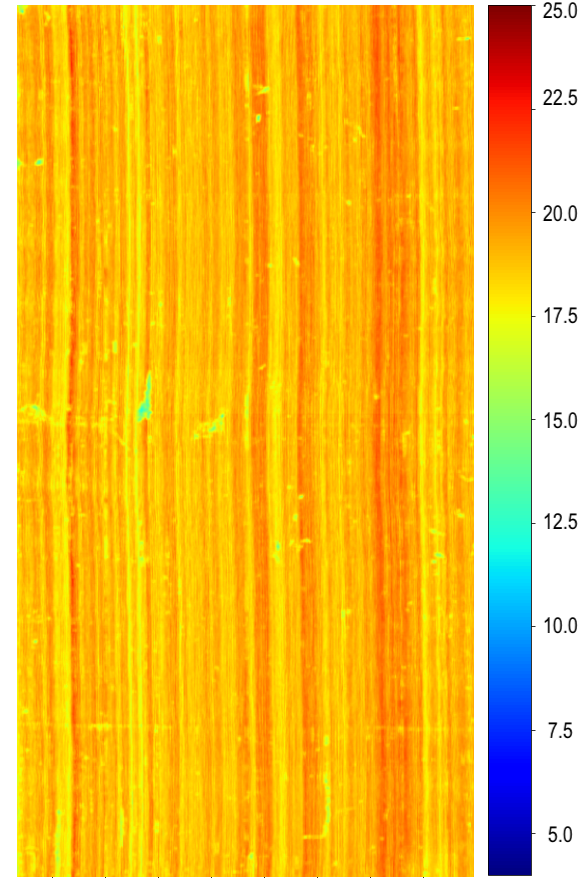} 
\\
 (a) & (b)
\end{tabular}
\caption{Vertical threads density maps for Diego del Corral y Arellano by Velázquez, \cite{P001195}: (a) using Segmentation, and (b) using Regression.} \label{fig:VerticalDM}
\end{figure}

We include the vertical density maps for Diego del Corral y Arellano estimated with segmentation DL, Fig. \ref{fig:VerticalDM}.a, and the proposed regression approach, in Fig. \ref{fig:VerticalDM}.b. The density map obtained using regression is cleaner, some noisy areas disappear and the vertical patterns have more continuity throughout the density map. 

\begin{figure}[!tp]
\centering
 \includegraphics[width=7cm]{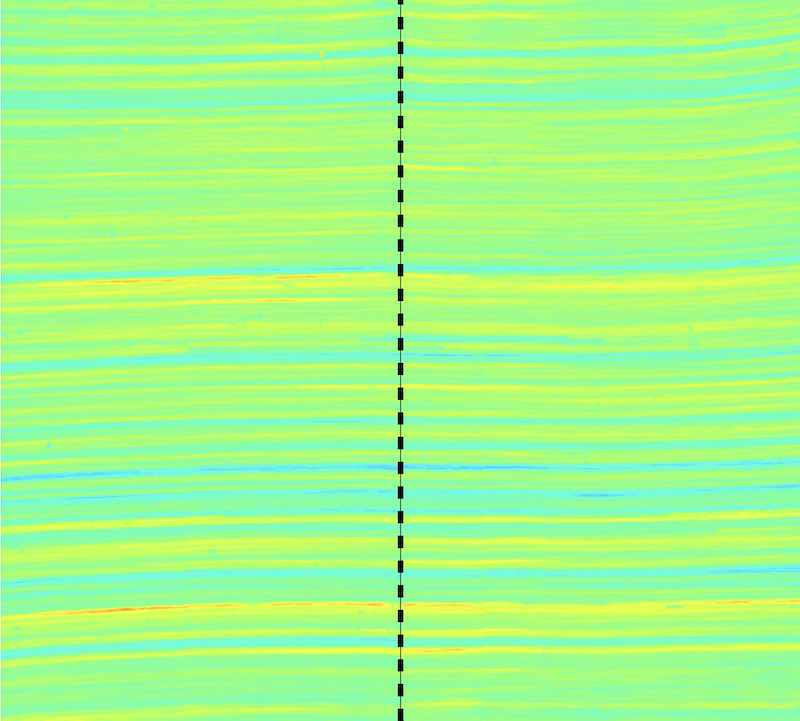}
\caption{Match of horizontal thread densities using Regression, in thr/cm, for Antonia de Ipeñarrieta and Son, to the left and Diego del Corral y Arellano, flipped horizontally, to the right.} \label{fig:Match}
\end{figure}

Finally, it can be clearly observed how the pattern of variations of the separation in the horizontal threads, computed with the novel regression DL method perfectly matches in both canvases (see Fig. \ref{fig:Match}). The image to the left corresponds to the horizontal density map of P001196 while the image to the right of the dashed line to the one of P001195, after a horizontal flip. Therefore, it can be concluded that both fabrics come from the same bolt. 

\subsection{Changing Authorship}

\begin{figure}[!tp]
\centering
\begin{tabular}{cccc}
 \includegraphics[width=4cm]{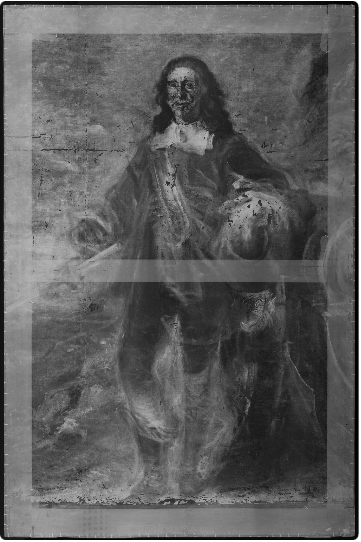}& 
  \includegraphics[width=3.2cm]{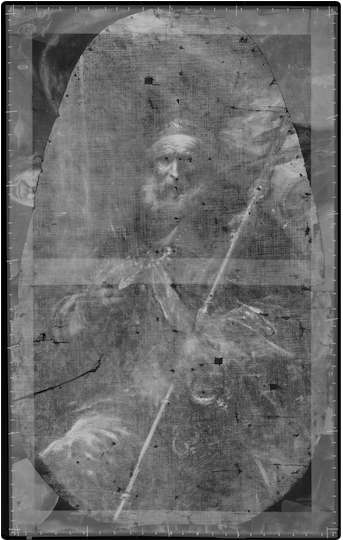}\\
(a) & (b) 
\end{tabular}
\caption{Paintings (a) An Artillery General by Rizi \cite{P001127} and (b)  Pope St. Leo I the Great  \cite{P007113} by Herrera El Mozo.} \label{fig:RizziHerX}
\end{figure}

\begin{figure*}[!tp]
\centering
\begin{tabular}{cccc}
\includegraphics[width=16cm]{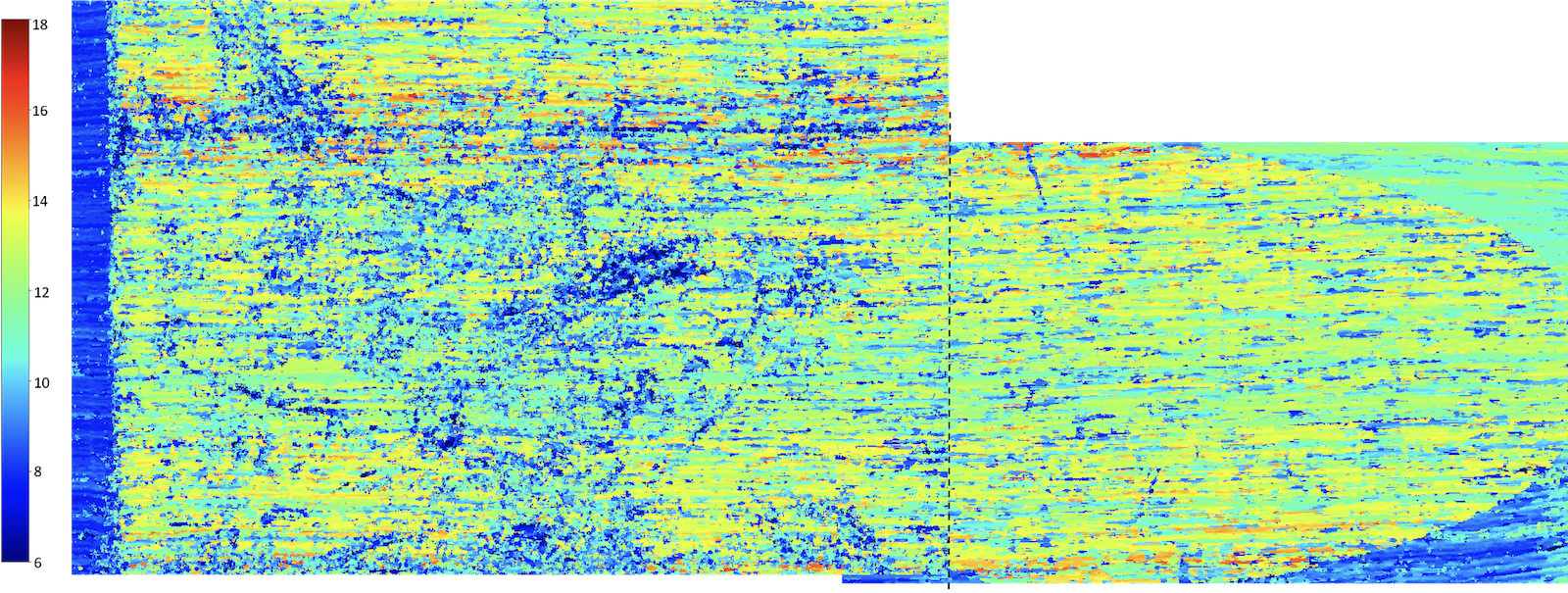} \\
 (a) \\
 \includegraphics[width=16cm]{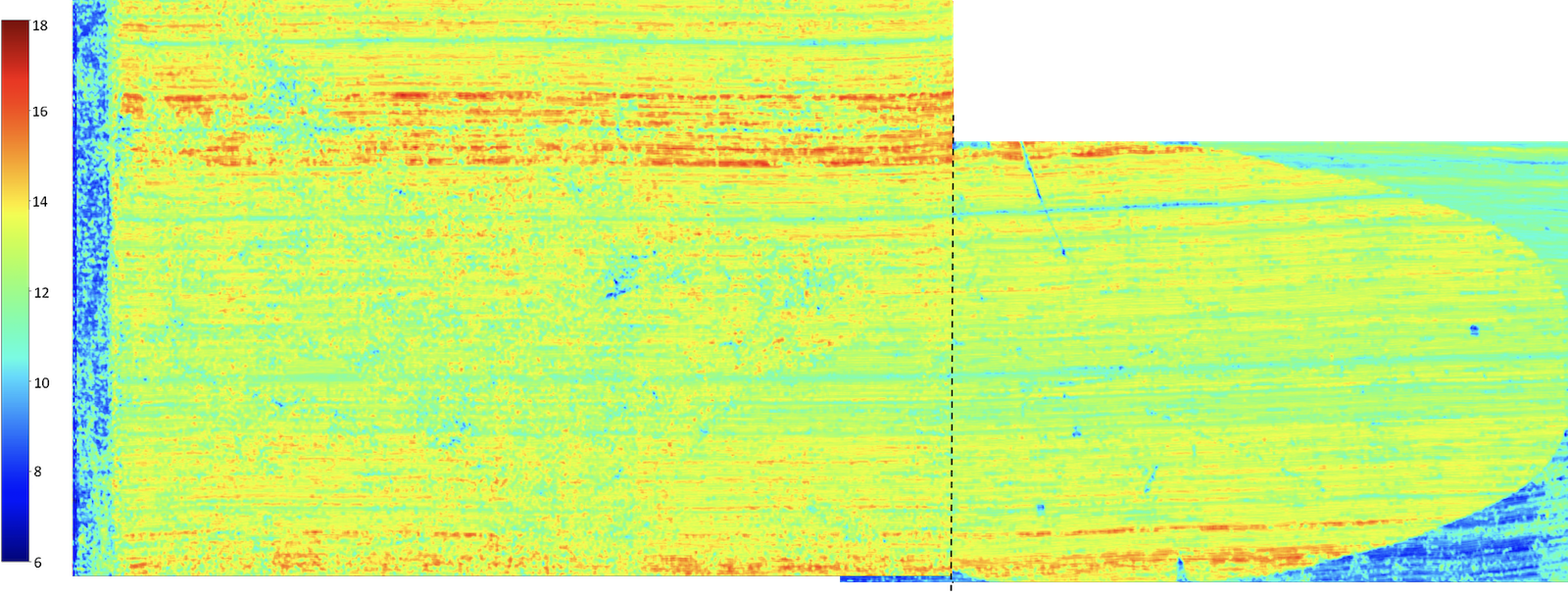}\\
 (b) \\
\end{tabular}
\caption{Match of vertical thread densities in thr/cm for P001127 and P007113 using (a) FT and (b) Regression DL. Canvases have been rotated 90$^\circ$ clockwise to easy the representation. Also, the canvas P001127 has been overlapped over P007113 to better observe the matching.} \label{fig:Rizi}
\end{figure*}

In this case, we process a pair of two canvases at the Museo N. del Prado,  An Artillery General  (P001127) \cite{P001127} and Pope St. Leo I the Great (P007113) \cite{P007113} whose authorship had been attributed to Francisco Rizi and Francisco de Herrera El Mozo, respectively. The X-ray images can be observed in Fig. \ref{fig:RizziHerX}. At the Dep. of Technical Documentation of the Museo N. del Prado the curators had some doubts about the authorship of the first one and faced the study of both paintings under the hypothesis that the first one could have been painted by Herrera El Joven. Within the studies performed, the analysis of the fabric was included. A first analysis was performed with FT and the Aracne software \cite{Murillo14,Alba21}, obtaining inconclusive results, as the FT provided quite noisy outcomes. An FT-based study is included in Fig. \ref{fig:Rizi}.a.
It can be observed that the FT analysis is quite noisy and it is quite complicated to conclude about the matching. By using the new inception regression DL approach we were able to match the vertical thread densities, concluding that the fabrics came from the same bolt. The matching result is included in Fig. \ref{fig:Rizi}.b. This result was used to arrange the FT threads density maps in Fig. \ref{fig:Rizi}.a.
%
The matching found between the fabrics of these two works helped El Museo N. del Prado to change the authorship of the An Artillery General  (P001127), \cite{P001127} now attributed to Herrera El Mozo, starting April 2023.

\subsection{Improvement in Execution Time}
By using the regression DL approach we avoid the signal processing algorithm to translate the segmentation into thread densities. Hence, we obtain a reduction in the running time. 
In Table \ref{tab:comptimesPlates} we report the times needed (in hours) to process some of the X-ray plates in the previous studies. We include the times invested by segmentation, regression and SS regression DL approaches. 

\begin{table}[htbp]
\centering
\caption{Execution time, in hours, to obtain the thread density maps of canvases.}\label{tab:comptimesPlates}
\begin{tabular}{c | c c c c c c c}
\toprule
    Plate & P1195 & P1114 & P1127 & NG6477 \\
\midrule
Segmentation & 2.51 & 1.51 & 5.44 & 19.53 \\
\hline
Regression & 2.01 & 1.06 & 4.72 & 16.51 \\
\hline
SS Regression & 5.76 & 3.77 & 12.79 & 44.95 \\
\bottomrule
\end{tabular}
\end{table}

As can be observed in Table \ref{tab:comptimesPlates}, the regression method is faster than the previous segmentation DL approach, since it avoids using the referred processing algorithms. The improvement is especially noteworthy in plates where a larger number of patches are processed (P001127 and NG6477). It can also be verified how the execution of the SS learning algorithm involves a relevant increase in the running time since it requires generating the samples and re-training. It can be concluded that SS learning is recommended if we can obtain a noticeable improvement, especially in those where FT offers good results. 

Finally, we have also found that the regression approach is more efficient in managing memory resources. Although the regression model has more parameters than its segmentation counterpart, using the regression approach avoids having to manage a segmentation map for each patch. The difference between storing one model or another in RAM is not relevant, since the difference is 63MB in favor of the segmentation model. However, the difference between handling and not handling the segmentation maps is in the order of several GB, depending on the size of the plate and how it is programmed. Note that we need to estimate the output images of the segmentation, and if parallel programming is exploited this involves handling several 40,000 pixels images, typically thousands of them. Therefore, the better and more efficient performance of the regression model is justified, both in execution time and in memory management.


%

\section{Conclusions}


In the novel approach presented, we resort to the regression DL paradigm to directly estimate the thread density. We avoid the signal processing stage of previous segmentation DL methods and are able to define a loss function to measure the error in thread counting at the output. Several models have been proposed, where hyperparameters have been selected through optimized search. 
Transfer learning from the segmentation DL approach did not provide good results. Lower errors were achieved by learning the encoder and the dense layers from scratch. Including residual connections did not further reduce the error, while by exploiting the VGG model we achieved slightly better results. 

Another line of research in this work was focused on pre-processing and DA. We proposed to increase the data set by including central patches from the labeled samples. Also, we reduce the maximum allowed rotations in the DA. On the other hand, in the pre-processing, we include histogram equalization and variable window size filtering. An ablation study is included to underline the benefits of these improvements. These methods can be also successfully applied to the segmentation DL approach.

The labeling of samples is a tedious task that limits the performance of the DL approaches. We propose a SS algorithm based on generating labels from similar estimations for the FT and the regression DL. With this method, other models with a larger need for samples could be explored \cite{Gao2023}. In the test dataset, we further reduce the NMAE from 1.01\% to 0.94\%. Compared to the FT (7.47\%), or the segmentation DL (1.61\%), we achieve a remarkable reduction. Running times are also improved by about 20\%. In summary, by directly estimating the thread densities we not only reduce running times but we can minimize the error itself in the training of the model, quite reducing it. 





Several studies have been proposed to illustrate the good performance of our new approach. First, in favorable scenarios to the FT such as the analysis of Ixion by Ribera, the SS regression DL exhibits a good unblurred, and noiseless result. In the study of the paintings by Poussin we have evidenced that DL can be used as an ML tool with no need for extra-labeling. Furthermore, the presented tool has a good performance in the estimations of densities in the whole needed range while the ATCA has a noisy behavior for low values of thread densities. It is interesting to note that ATCA has a good spatial resolution. A redesigning of the regression DL approach to improve spatial resolution remains a future line of research. When analyzing Velazquez, where the FT fails, the regression DL provides even better results compared to the segmentation DL algorithm. Finally, we face a recent case study in El Museo N. del Prado, where the method helped to conclude the change of authorship of a masterwork by Rizi, now attributed to Herrera el Joven.

\section*{Code}
Code in python is available at \url{https://github.com/gapsc-us/DL4ART}. There you will find  1) a sample of input image and label, we used labelme to annotate the vertical and horizontal threads, 2) a .py with the models used, and 3) a jupyter notebook to train the regression VGG DL model using a GPU.
We cannot share or distribute images of the canvases, just a sample is included with the code. 

%

\section*{Acknowledgements}
We are really grateful to The National Gallery and in particular to Dr. Catherine Higgitt for facilitating the X-ray plates of the canvases by N. Poussin. 

\section*{Funding}
{This work was supported by  the Consejería de Economía y Conocimiento, Junta de Andalucía and FEDER-European Union [{ATENEA}, P20\_01216], 
 and Ministerio de Ciencia e Innovación de España and FEDER-European Union [EPiCENTER, PID2021-123182OB-I00], [MCIN/AEI/10.13039/501100011033], [Skin, PID2021-127871OB-I00].}


  \bibliographystyle{elsarticle-num-names}


\begin{thebibliography}{34}
\expandafter\ifx\csname natexlab\endcsname\relax\def\natexlab#1{#1}\fi
\providecommand{\url}[1]{\texttt{#1}}
\providecommand{\href}[2]{#2}
\providecommand{\path}[1]{#1}
\providecommand{\DOIprefix}{doi:}
\providecommand{\ArXivprefix}{arXiv:}
\providecommand{\URLprefix}{URL: }
\providecommand{\Pubmedprefix}{pmid:}
\providecommand{\doi}[1]{\href{http://dx.doi.org/#1}{\path{#1}}}
\providecommand{\Pubmed}[1]{\href{pmid:#1}{\path{#1}}}
\providecommand{\bibinfo}[2]{#2}
\ifx\xfnm\relax \def\xfnm[#1]{\unskip,\space#1}\fi
\bibitem[{Alba-Carcel{\'e}n and Murillo-Fuentes(2021)}]{Alba21}
\bibinfo{author}{L.~Alba-Carcel{\'e}n}, \bibinfo{author}{J.~J.
  Murillo-Fuentes},
\newblock \bibinfo{title}{Fabrics as a painting support. new tools for the
  study},
\newblock in: \bibinfo{booktitle}{{La ciencia y el arte. Ciencias
  experimentales y conservaci{\'o}n del patrimonio}},
  \bibinfo{publisher}{Ministerio de Cultura y Deporte}, \bibinfo{year}{2021},
  pp. \bibinfo{pages}{219--230}. \DOIprefix\doi{10.1007/978-3-319-75316-4_7}.
\bibitem[{de~Carbonnel(1980)}]{Vanderlip80}
\bibinfo{author}{K.~V. de~Carbonnel},
\newblock \bibinfo{title}{A study of french painting canvases},
\newblock \bibinfo{journal}{Journal of the American Institute for Conservation}
  \bibinfo{volume}{20} (\bibinfo{year}{1980}) \bibinfo{pages}{3--20}.
\bibitem[{Simois and Murillo-Fuentes(2018)}]{Simois18}
\bibinfo{author}{F.~J. Simois}, \bibinfo{author}{J.~J. Murillo-Fuentes},
\newblock \bibinfo{title}{On the power spectral density applied to the analysis
  of old canvases},
\newblock \bibinfo{journal}{Signal Processing} \bibinfo{volume}{143}
  (\bibinfo{year}{2018}) \bibinfo{pages}{253--268}.
\bibitem[{Barlow(1878)}]{Barlow1878}
\bibinfo{author}{A.~Barlow}, \bibinfo{title}{The History and Principles of
  Weaving by Hand and by Power}, \bibinfo{publisher}{Low, Marston, Searle and
  Rivington}, \bibinfo{year}{1878}.
\bibitem[{Johnson et~al.(2010)Johnson, Sun, Johnson, and
  Hendriks}]{Johnson2010}
\bibinfo{author}{D.~H. Johnson}, \bibinfo{author}{L.~Sun},
  \bibinfo{author}{C.~R. Johnson}, \bibinfo{author}{E.~Hendriks},
\newblock \bibinfo{title}{Matching canvas weave patterns from processing
  {X}-ray images of master paintings},
\newblock in: \bibinfo{booktitle}{IEEE International Conference on Acoustics,
  Speech and Signal Processing (ICASSP)}, \bibinfo{year}{2010}.
  \DOIprefix\doi{10.1109/ICASSP.2010.5495297}.
\bibitem[{Johnson et~al.(2013)Johnson, Jr., and Erdmann}]{Johnson2013}
\bibinfo{author}{D.~H. Johnson}, \bibinfo{author}{C.~R.~J. Jr.},
  \bibinfo{author}{R.~G. Erdmann},
\newblock \bibinfo{title}{Weave analysis of paintings on canvas from
  radiographs},
\newblock \bibinfo{journal}{Signal Processing} \bibinfo{volume}{93}
  (\bibinfo{year}{2013}) \bibinfo{pages}{527--540}.
\bibitem[{Rumelhart et~al.(1986)Rumelhart, Hinton, and Williams}]{AE86}
\bibinfo{author}{D.~Rumelhart}, \bibinfo{author}{G.~Hinton},
  \bibinfo{author}{R.~Williams}, \bibinfo{title}{Neurocomputing: foundations of
  research, learning internal representations by error propagation},
  \bibinfo{year}{1986}.
\bibitem[{Goodfellow et~al.(2016)Goodfellow, Bengio, and
  Courville}]{Goodfellow2016}
\bibinfo{author}{I.~Goodfellow}, \bibinfo{author}{Y.~Bengio},
  \bibinfo{author}{A.~Courville}, \bibinfo{title}{{Deep Learning}},
  \bibinfo{publisher}{MIT Press}, \bibinfo{year}{2016}.
\bibitem[{Bejarano et~al.(2022{\natexlab{a}})Bejarano, Murillo-Fuentes, and
  Alba-Carcel{\'e}n}]{Bejarano2022a}
\bibinfo{author}{A.~D. Bejarano}, \bibinfo{author}{J.~Murillo-Fuentes},
  \bibinfo{author}{L.~Alba-Carcel{\'e}n},
\newblock \bibinfo{title}{{C}rossings segmentation in plain weaves for {X}-rays
  of canvases with deep learning},
\newblock in: \bibinfo{booktitle}{7th IP4AI meeting Computational approaches
  for technical imaging in cultural heritage}, \bibinfo{organization}{National
  Gallery}, \bibinfo{year}{2022}{\natexlab{a}}.
\bibitem[{Bejarano et~al.(2022{\natexlab{b}})Bejarano, Murillo-Fuentes, and
  Alba-Carcel{\'e}n}]{Bejarano2022b}
\bibinfo{author}{A.~D. Bejarano}, \bibinfo{author}{J.~Murillo-Fuentes},
  \bibinfo{author}{L.~Alba-Carcel{\'e}n},
\newblock \bibinfo{title}{Crossings segmentation in plain weaves for {X}-rays
  of canvases with deep learning: technical details},
\newblock in: \bibinfo{booktitle}{7th IP4AI meeting Computational approaches
  for technical imaging in cultural heritage}, \bibinfo{organization}{National
  Gallery}, \bibinfo{year}{2022}{\natexlab{b}}.
\bibitem[{Delgado et~al.(2023)Delgado, Alba-Carcel{\'e}n, and
  Murillo-Fuentes}]{Bejarano2023}
\bibinfo{author}{A.~Delgado}, \bibinfo{author}{L.~Alba-Carcel{\'e}n},
  \bibinfo{author}{J.~Murillo-Fuentes}, \bibinfo{title}{Crossing points
  detection in plain weave for old paintings with deep learning},
  \bibinfo{howpublished}{http://arxvii.org/}, \bibinfo{year}{2023}.
\bibitem[{Shorten and Khoshgoftaar(2019)}]{DA2019}
\bibinfo{author}{C.~Shorten}, \bibinfo{author}{T.~M. Khoshgoftaar},
\newblock \bibinfo{title}{A survey on image data augmentation for deep
  learning},
\newblock \bibinfo{journal}{Journal of big data} \bibinfo{volume}{6}
  (\bibinfo{year}{2019}) \bibinfo{pages}{1--48}.
\bibitem[{Snoek et~al.(2012)Snoek, Larochelle, and Adams}]{Snoek12}
\bibinfo{author}{J.~Snoek}, \bibinfo{author}{H.~Larochelle},
  \bibinfo{author}{R.~P. Adams},
\newblock \bibinfo{title}{Practical bayesian optimization of machine learning
  algorithms},
\newblock in: \bibinfo{editor}{F.~Pereira}, \bibinfo{editor}{C.~Burges},
  \bibinfo{editor}{L.~Bottou}, \bibinfo{editor}{K.~Weinberger} (Eds.),
  \bibinfo{booktitle}{Advances in Neural Information Processing Systems},
  volume~\bibinfo{volume}{25}, \bibinfo{publisher}{Curran Associates, Inc.},
  \bibinfo{year}{2012}. \URLprefix
  \url{https://proceedings.neurips.cc/paper/2012/file/05311655a15b75fab86956663e1819cd-Paper.pdf}.
\bibitem[{O'Malley et~al.(2019)O'Malley, Bursztein, Long, Chollet, Jin,
  Invernizzi et~al.}]{Omalley19}
\bibinfo{author}{T.~O'Malley}, \bibinfo{author}{E.~Bursztein},
  \bibinfo{author}{J.~Long}, \bibinfo{author}{F.~Chollet},
  \bibinfo{author}{H.~Jin}, \bibinfo{author}{L.~Invernizzi}, et~al.,
  \bibinfo{title}{Kerasturner},
  \bibinfo{howpublished}{\url{https://github.com/keras-team/keras-tuner}},
  \bibinfo{year}{2019}.
\bibitem[{{Maaten} and Erdmann(2015)}]{Maaten15}
\bibinfo{author}{L.~{Maaten}}, \bibinfo{author}{R.~G. Erdmann},
\newblock \bibinfo{title}{{Automatic thread-level canvas analysis: A
  machine-learning approach to analyzing the canvas of paintings}},
\newblock \bibinfo{journal}{IEEE Signal Process. Mag.} \bibinfo{volume}{32}
  (\bibinfo{year}{2015}).
\bibitem[{Escofet et~al.(2001)Escofet, Mill{\'{a}}n, and
  Rall{\'{o}}}]{Escofet2001}
\bibinfo{author}{J.~Escofet}, \bibinfo{author}{M.~S. Mill{\'{a}}n},
  \bibinfo{author}{M.~Rall{\'{o}}},
\newblock \bibinfo{title}{{Modeling of woven fabric structures based on Fourier
  image analysis}},
\newblock \bibinfo{journal}{Applied Optics}  (\bibinfo{year}{2001}).
\bibitem[{Fond\'on-Garc\'ia et~al.(2014)Fond\'on-Garc\'ia, Simois, and
  Murillo-Fuentes}]{Murillo14}
\bibinfo{author}{I.~Fond\'on-Garc\'ia}, \bibinfo{author}{F.~J. Simois},
  \bibinfo{author}{J.~J. Murillo-Fuentes},
\newblock \bibinfo{title}{Software tool for thread counting in {X}-rays of
  plain-weave painting canvas},
\newblock in: \bibinfo{booktitle}{International Conference on Non-Destructive
  Investigations and Microanalysis for the Diagnostics and Conservation of
  Cultural and Environmental Heritage ({ART2014})}, \bibinfo{address}{Madrid,
  Spain}, \bibinfo{year}{2014}, p. \bibinfo{pages}{IND119}.
\bibitem[{Sizyakin et~al.(2020)Sizyakin, Cornelis, Meeus, Dubois, Martens,
  Voronin, and Pizurica}]{Sizyakin20}
\bibinfo{author}{R.~Sizyakin}, \bibinfo{author}{B.~Cornelis},
  \bibinfo{author}{L.~Meeus}, \bibinfo{author}{H.~Dubois},
  \bibinfo{author}{M.~Martens}, \bibinfo{author}{V.~Voronin},
  \bibinfo{author}{A.~Pizurica},
\newblock \bibinfo{title}{Crack detection in paintings using convolutional
  neural networks},
\newblock \bibinfo{journal}{IEEE Access} \bibinfo{volume}{8}
  (\bibinfo{year}{2020}) \bibinfo{pages}{74535 -- 74552}. \bibinfo{note}{Cited
  by: 8; All Open Access, Gold Open Access, Green Open Access}.
\bibitem[{Roberto et~al.(2020)Roberto, Ortego, and Davis}]{Roberto2020}
\bibinfo{author}{J.~Roberto}, \bibinfo{author}{D.~Ortego},
  \bibinfo{author}{B.~Davis},
\newblock \bibinfo{title}{Toward the automatic retrieval and annotation of
  outsider art images: A preliminary statement},
\newblock in: \bibinfo{booktitle}{AI4HI}, \bibinfo{year}{2020}.
\bibitem[{{Pu} et~al.(2020){Pu}, {Sober}, {Daly}, {Higgitt}, {Daubechies}, and
  {Rodrigues}}]{Pu2020}
\bibinfo{author}{W.~{Pu}}, \bibinfo{author}{B.~{Sober}},
  \bibinfo{author}{N.~{Daly}}, \bibinfo{author}{C.~{Higgitt}},
  \bibinfo{author}{I.~{Daubechies}}, \bibinfo{author}{M.~R.~D. {Rodrigues}},
\newblock \bibinfo{title}{A connected auto-encoders based approach for image
  separation with side information: With applications to art investigation},
\newblock in: \bibinfo{booktitle}{IEEE Int. Conf. on Acoustics, Speech and
  Signal Process. (ICASSP)}, \bibinfo{year}{2020}, pp.
  \bibinfo{pages}{2213--2217}.
\bibitem[{Zou et~al.(2021)Zou, Zhao, and Zhao}]{Zou21}
\bibinfo{author}{Z.~Zou}, \bibinfo{author}{P.~Zhao}, \bibinfo{author}{X.~Zhao},
\newblock \bibinfo{title}{Virtual restoration of the colored paintings on
  weathered beams in the forbidden city using multiple deep learning
  algorithms},
\newblock \bibinfo{journal}{Advanced Engineering Informatics}
  \bibinfo{volume}{50} (\bibinfo{year}{2021}). \bibinfo{note}{Cited by: 0}.
\bibitem[{Ronneberger et~al.(2015)Ronneberger, Fischer, and Brox}]{Unet15}
\bibinfo{author}{O.~Ronneberger}, \bibinfo{author}{P.~Fischer},
  \bibinfo{author}{T.~Brox},
\newblock \bibinfo{title}{{U-N}et: Convolutional networks for biomedical image
  segmentation},
\newblock in: \bibinfo{booktitle}{Lecture Notes in Computer Science},
  Artificial Intelligence and Lecture Notes in Bioinformatics,
  \bibinfo{year}{2015}. \DOIprefix\doi{10.1007/978-3-319-24574-4_28}.
\bibitem[{Simonyan and Zisserman(2015)}]{Simonyan2015}
\bibinfo{author}{K.~Simonyan}, \bibinfo{author}{A.~Zisserman},
\newblock \bibinfo{title}{Very deep convolutional networks for large-scale
  image recognition},
\newblock \bibinfo{publisher}{Computational and Biological Learning Society},
  \bibinfo{year}{2015}, pp. \bibinfo{pages}{1--14}.
\bibitem[{Kingma and Ba(2015)}]{Kingma15}
\bibinfo{author}{D.~P. Kingma}, \bibinfo{author}{J.~Ba},
\newblock \bibinfo{title}{Adam: {A} method for stochastic optimization},
\newblock in: \bibinfo{booktitle}{3rd International Conference on Learning
  Representations, {ICLR} 2015, San Diego, CA, USA, May 7-9, 2015, Conference
  Track Proceedings}, \bibinfo{address}{San Diego, California, USA},
  \bibinfo{year}{2015}. \URLprefix \url{http://arxiv.org/abs/1412.6980}.
\bibitem[{Aradillas et~al.(2021)Aradillas, Murillo-Fuentes, and
  Olmos}]{Aradillas21}
\bibinfo{author}{J.~C. Aradillas}, \bibinfo{author}{J.~J. Murillo-Fuentes},
  \bibinfo{author}{P.~M. Olmos},
\newblock \bibinfo{title}{Boosting offline handwritten text recognition in
  historical documents with few labeled lines},
\newblock \bibinfo{journal}{IEEE Access} \bibinfo{volume}{9}
  (\bibinfo{year}{2021}) \bibinfo{pages}{76674--76688}.
\bibitem[{de~Ribera(1632)}]{P001114}
\bibinfo{author}{J.~de~Ribera}, \bibinfo{title}{Ixion},
  \bibinfo{howpublished}{Museo {N}acional del {P}rado (P001114)},
  \bibinfo{year}{1632}.
\bibitem[{Poussin(1636{\natexlab{a}})}]{NG6477}
\bibinfo{author}{N.~Poussin}, \bibinfo{title}{The {T}riumph of {P}an},
  \bibinfo{howpublished}{National Gallery (NG6477)},
  \bibinfo{year}{1636}{\natexlab{a}}.
\bibitem[{Poussin(1636{\natexlab{b}})}]{NG42}
\bibinfo{author}{N.~Poussin}, \bibinfo{title}{The {T}riumph of {S}ilenus},
  \bibinfo{howpublished}{National Gallery (NG6477)},
  \bibinfo{year}{1636}{\natexlab{b}}.
\bibitem[{{Maaten}(2015)}]{Maaten15b}
\bibinfo{author}{L.~{Maaten}}, \bibinfo{title}{Canvas analysis},
  \bibinfo{year}{2015}. \URLprefix \url{https://lvdmaaten.github.io/canvas/}.
\bibitem[{de~Silva~y Vel{\'a}zquez(1632{\natexlab{a}})}]{P001196}
\bibinfo{author}{D.~R. de~Silva~y Vel{\'a}zquez}, \bibinfo{title}{Antonia de
  {I}pe{\~n}arrieta y {G}ald{\'o}s and her {S}on, {L}uis},
  \bibinfo{howpublished}{Museo Nacional del Prado (P001196)},
  \bibinfo{year}{1632}{\natexlab{a}}.
\bibitem[{de~Silva~y Vel{\'a}zquez(1632{\natexlab{b}})}]{P001195}
\bibinfo{author}{D.~R. de~Silva~y Vel{\'a}zquez}, \bibinfo{title}{Diego del
  {C}orral y {A}rellano}, \bibinfo{howpublished}{Museo Nacional del Prado
  (P001195)}, \bibinfo{year}{1632}{\natexlab{b}}.
\bibitem[{Rizi(1660)}]{P001127}
\bibinfo{author}{F.~Rizi}, \bibinfo{title}{An {A}rtillery {G}eneral},
  \bibinfo{howpublished}{Museo Nacional del Prado (P001127)},
  \bibinfo{year}{1660}.
\bibitem[{Mozo(tury)}]{P007113}
\bibinfo{author}{F.~D. H.~E. Mozo}, \bibinfo{title}{Pope {S}t. {L}eo {I} the
  {G}reat}, \bibinfo{howpublished}{Museo Nacional del Prado (P007113)},
  \bibinfo{year}{17th Century}.
\bibitem[{Gao et~al.(2023)Gao, Yang, Zhang, Goulermas, Geng, Yan, and
  Huang}]{Gao2023}
\bibinfo{author}{P.~Gao}, \bibinfo{author}{X.~Yang},
  \bibinfo{author}{R.~Zhang}, \bibinfo{author}{J.~Y. Goulermas},
  \bibinfo{author}{Y.~Geng}, \bibinfo{author}{Y.~Yan},
  \bibinfo{author}{K.~Huang},
\newblock \bibinfo{title}{Generalized image outpainting with u-transformer},
\newblock \bibinfo{journal}{Neural Networks} \bibinfo{volume}{162}
  (\bibinfo{year}{2023}) \bibinfo{pages}{1--10}.

\end{thebibliography}

\end{document}